%% file: main.tex
\documentclass{article} % For LaTeX2e
\usepackage[preprint]{colm2026_conference}

\usepackage{microtype}
\usepackage{hyperref}
\usepackage{url}
\usepackage{booktabs}
\usepackage{amsmath}
\usepackage{amssymb}
\usepackage{mathtools}
\usepackage{amsthm}
\usepackage{algorithm}
\usepackage{algpseudocode}
\usepackage{multirow}
\usepackage{graphicx}
\usepackage[table]{xcolor}
\usepackage{subcaption}
\usepackage{wrapfig} 
\usepackage{mdframed}
\usepackage{listings}
\usepackage{placeins}

\usepackage{caption}
\newmdenv[
  backgroundcolor=gray!5,
  linecolor=gray!40,
  roundcorner=3pt,
  skipabove=6pt,
  skipbelow=6pt
]{promptframe}

\theoremstyle{plain}
\newtheorem{theorem}{Theorem}[section]
\newtheorem{proposition}[theorem]{Proposition}

\theoremstyle{definition}

\theoremstyle{remark}

\definecolor{BaseRow}{HTML}{e8e8e8}
\definecolor{AllRow}{HTML}{E8F9FC}
\definecolor{BaselineRow}{HTML}{D4F0FA}
\definecolor{OursRow}{HTML}{9CE1F7}

\usepackage{lineno}

\definecolor{darkblue}{rgb}{0, 0, 0.5}
\hypersetup{colorlinks=true, citecolor=darkblue, linkcolor=darkblue, urlcolor=darkblue}

\title{Learn Less, Evolve More: Data Selection for Self-Improving LLM Agents}

\author{Ting-Wei Li,\ \ Sirui Chen,\ \ Jiaru Zou,\ \ Yingbing Huang, Tianxin Wei, \\ {\bf  Jingrui He,\ \ Hanghang Tong} \\
University of Illinois Urbana-Champaign\\
\small{
\texttt{\{twli, sirui6, jiaruz2, yh21, twei10, jingrui, htong\}@illinois.edu}
}
}

\newcommand{\method}{\textsc{EvoSelect}}

\begin{document}

\ifcolmsubmission
\linenumbers
\fi

\maketitle

\begin{abstract}
% \hh{the title is a bit long. consider these options: (1) drop 'via ...' part. (2) evolution for targeted task adaptation, isn't this exactly adapation? so maybe: EvoSelect: Data Efficient LLM Adaptation?}\tw{modified}

% \sr{I feel like the motivation and problem are not clear enough. Maybe we say the problem is from training on synthetic data (quality, noise, etc,) and then propose to address it through an iterative generation–training loop?}
% \tw{i now position the iterative generation-training loop as our setting instead of a solution (see Sec 2.3); or we just directly say this setting here, and say the generation-selection-training loop can solve the issue?}\sr{Yeah, I think this would be better. And we could also emphasize the concept of the "evolution" in this loop.}\tw{sure, will modify}

Adapting large language models (LLMs) to a targeted task efficiently and effectively remains a fundamental challenge. Such adaptation often requires iteratively improving the model toward a targeted task, yet collecting high-quality human-labeled data to support this process is costly and difficult to scale. As a result, synthetic data generation has emerged as a flexible and scalable alternative. One straightforward approach is through an \textit{iterative generation–training loop}, where candidate data are synthesized through an external generator, the model is updated using these data and the process is repeated over iterations. However, generated samples can be noisy, highly redundant, or even misaligned with the targeted task distribution. Training indiscriminately on such data can dilute useful learning signals and even degrade model performance. 
To address this, we introduce a refined paradigm, namely an \textit{iterative generation–selection–training loop}, which incorporates a selection step prior to model updates. Building on this paradigm, we propose \method, a data-efficient framework to evolve LLM effectively. Given candidate samples produced by the data generator, \method\ selects training data by jointly modeling targeted task alignment and diversity. We estimate task relevance through optimal transport with proxy gradient representations, which quantifies how well candidate samples align with the targeted task distribution. To mitigate redundancy, we incorporate a diversification mechanism that promotes coverage of complementary training samples. By interleaving alignment and diversification, \method\ enables progressive LLM evolution toward targeted tasks. Extensive experiments on various benchmarks demonstrate that with either weak or strong data generators, \method\ consistently improves adaptation efficacy over existing data selection methods.
\end{abstract}

\input{sections/01Intro}
\input{sections/02Prelim}

\input{sections/03Method}

\input{sections/04Experiments}

\input{sections/05Related}
\input{sections/06Conclusion}

% \section*{Author Contributions}

% \section*{Acknowledgments}
% Use unnumbered first level headings for the acknowledgments. All

% \section*{Ethics Statement}
% Authors can add an optional ethics statement to the paper. 
% For papers that touch on ethical issues, this section will be evaluated as part of the review process. The ethics statement should come at the end of the paper. It does not count toward the page limit, but should not be more than 1 page. 

\bibliography{colm2026_conference}
\bibliographystyle{colm2026_conference}

\newpage
\appendix

\input{sections/99Appendix}

\end{document}

%% file: sections/01Intro.tex
\section{Introduction}

\begin{wrapfigure}{r}{0.6\textwidth}
\setlength{\belowcaptionskip}{-10pt}
\centering
\includegraphics[width=0.58\textwidth]{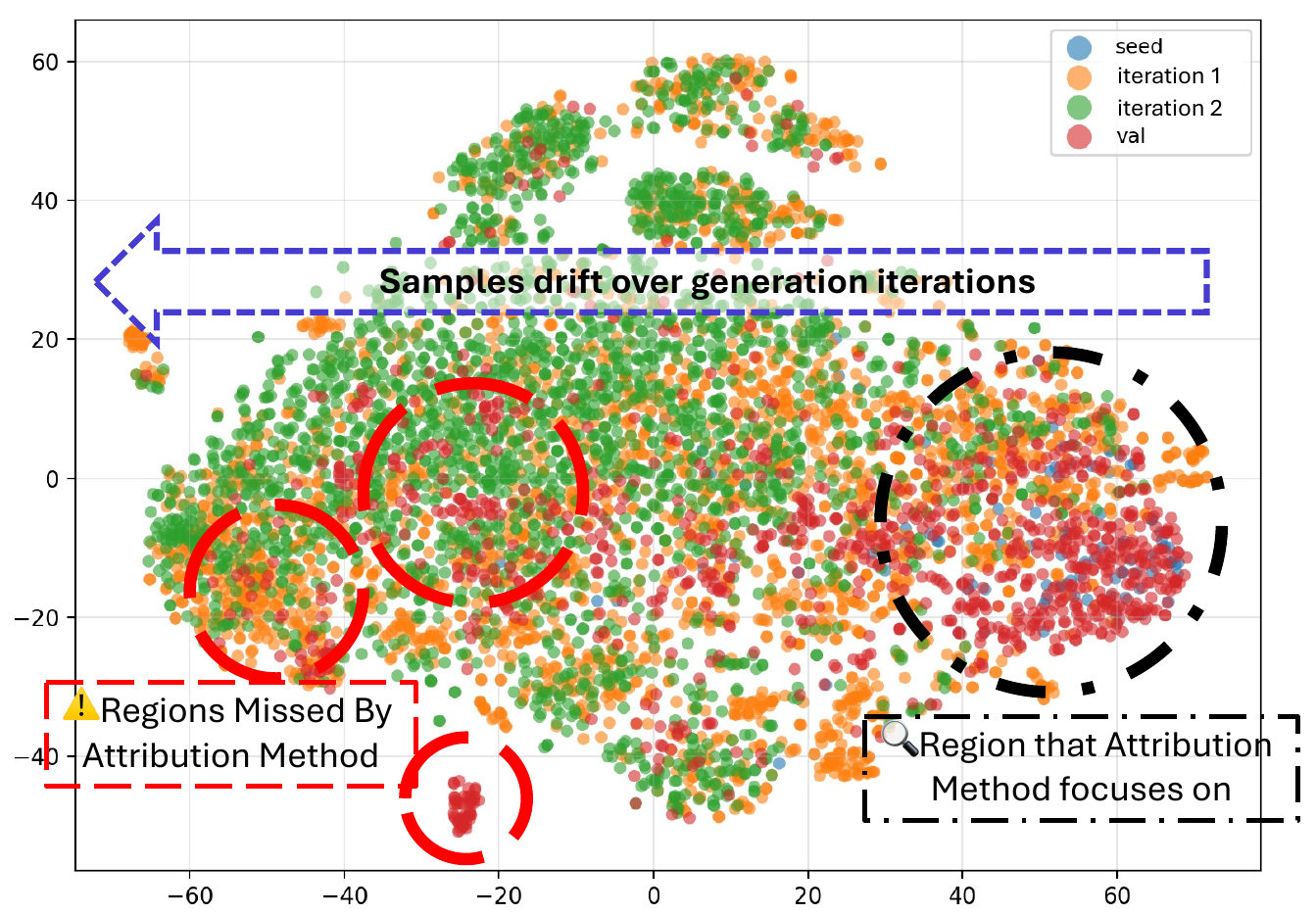}
  \caption{Issues of synthetic training samples and naive targeted selection. (1) Generated samples (\textcolor{orange}{\textit{orange}} \& \textcolor{teal}{\textit{green}}) drift away from the target distribution (\textcolor{purple}{\textit{red}}); (2) Generated samples increasingly overlap across iterations; (3) Attribution-only methods bias toward the centroid and under-cover the entire target distribution.
  % \sr{Is this the first place we use the term "attribution methods"?}\tw{we will refer this figure again after we intro attribution method in the next paragraph}
  }
\label{fig:intro-fig}
\end{wrapfigure}

Large language models (LLMs) have demonstrated strong general-purpose capabilities and are also widely deployed as powerful autonomous agents~\citep{minaee2024large,wei2026agentic}. However, many real-world applications require adapting LLMs to specific task distributions under limited supervision and constrained post-training resources~\citep{zhang2026instruction,han2024parameter,zhang2025survey, albalak2024survey}. To overcome the limitation, one prevalent solution is via a practical setting: \textit{iterative generation--training loop}~\citep{zelikman2022star,
singh2023beyond,yang2025spend}, where candidate data is synthesized by an external data generator, used to update the base model through fine-tuning, and the process is repeated over multiple iterations. 
% In these settings, the central challenge is two-fold. 
% To begin with, the target task suffers from severe \textit{data scarcity}, where high-quality labeled data are limited. This necessitates the use of synthetic data as a scalable alternative, which has been widely adopted in supervised fine-tuning pipelines~\citep{long2024llms,wang2024survey}.
However, even with abundant synthetic data, new challenges emerge: \textit{data efficacy} and \textit{data efficiency}. Firstly, not all generated examples are equally useful.
Under the iterative setting, these generated samples can progressively drift away from the targeted task, leading to reduced effectiveness for adaptation. Secondly, even well-aligned samples may become redundant over successive iterations, adding little new information while increasing unnecessary computational cost. We demonstrate these issues in Figure~\ref{fig:intro-fig}, where we visualize validation and synthetic training samples generated in different iterations with t-SNE on the \texttt{MMLU} dataset~\citep{hendrycks2020measuring}. We can observe that newly generated samples exhibit two key behaviors over iterations: (1) progressively mis-aligned with the target distribution (i.e. the \textcolor{blue}{blue} arrow: iteration 2 generated samples (\textcolor{teal}{\textit{green}}) move further away from the target distribution (\textcolor{purple}{\textit{red}}) compared to iteration 1 generated samples (\textcolor{orange}{\textit{orange}})) and (2) substantial overlap between successive iterations (i.e. \textcolor{orange}{\textit{orange}} and \textcolor{teal}{\textit{green}} samples). These observations highlight the risk that naively training on all generated data can dilute useful information and even degrade adaptation performance.
% \sr{What is new-batch? how can we derive these two observations from the fig? And what is the relation between the two key points here and the highlights in the fig?}\tw{it is "iteration" (fixed), green/red drift away from red (added arrow), highlight is for the issue in the next paragraph}\sr{Looks better now.} 

Thus, we are motivated to study a novel paradigm about effective LLM evolution for targeted task adaptation: an \textit{iterative generation-selection-training} loop, which selects high-quality data before updating the base model. Accordingly, the key question now becomes: \textbf{which generated examples should be used within the iterative loop?} 

% \tw{As demonstrated in Figure xxx and Table yyy, we can observe that ....}\tw{one attribution score dropping, one tsne plot to show overlapping}

% \setlength{\columnsep}{10pt}
% \begin{wraptable}{r}{0.6\textwidth} % r = right, l = left
% % \setlength{\tabcolsep}{6pt} 
% \centering
% \begin{tabular}{cccc}
% \toprule
% \midrule
%  & \textbf{ARC-C} & \textbf{CsQA} & \textbf{HeadQA} \\
% \midrule
% Mean. Attr. &  &  &  \\
% Mean. Cos. &  &  &  \\
% Perf. &  &  &  \\
%  \midrule
% \bottomrule
% \end{tabular}
% \caption{use all data performance drop}
% \label{tab:intro_figure}
% \end{wraptable}

% \sr{Can we say, on top of that, even we have large aligned data, many of them are redundant ...?} \tw{modified}\hh{does redundancy also wasts the training time?}\tw{modified}

% \setlength{\columnsep}{10pt}
% \begin{wraptable}{r}{0.6\textwidth} % r = right, l = left
% % \setlength{\tabcolsep}{6pt} 
% \centering
% \begin{tabular}{cccc}
% \toprule
% \midrule
%  & \textbf{ARC-C} & \textbf{CsQA} & \textbf{HeadQA} \\
% \midrule
% Mean. Attr. &  &  &  \\
% Mean. Cos. &  &  &  \\
% Perf. &  &  &  \\
%  \midrule
% \bottomrule
% \end{tabular}
% \caption{use all data performance drop}
% \label{tab:intro_figure}
% \end{wraptable}

To solve this puzzle, we argue that effective data selection in an iterative setting must jointly consider two objectives across iterations. On one hand, selected examples should be \emph{target-aligned}, meaning they need to stay functionally relevant to the targeted tasks. On the other hand, selected examples must maintain \emph{diversity}, covering complementary regions of the data space to avoid accumulating redundancy over iterations. However, existing approaches typically address these objectives in isolation. Some methods leverage training data attribution~\citep{deng2025survey} to prioritize target-aligned examples~\citep{xia2024less,zhou2024hyperinf,kwon2023datainf}, while others emphasize diversity through naive sampling or clustering strategies~\citep{ding2023enhancing,song2024scaling,ling2025diversity}. Even when both aspects are considered, they are often combined in a sequential/heuristic manner, such as through weighted sums, multiplication or local diversity adjustment~\citep{liu2024tsds,wu2024best,zhang2025d3}.

% As a result, these approaches either favor highly similar examples that lack diversity or introduce diverse but weakly relevant samples, leading to suboptimal training sets in our studied \textit{iterative generation-selection-training} setting. \sr{I think the discussion above reads more like a general data selection issue rather than a problem specific to our generation–training setting. Could we make it more setting-specific? We should also consider putting more emphasis on the notion of evolution here.}\tw{i  tweak the tone and make connection to the figure 1...let me know if still shallow}\sr{Can we claim that without a principled selection mechanism, the issue illustrated in Fig. 1 could compound over successive iterations?}

% data-attribution-based approaches  rely on data attribution to estimate example utility [cite], where samples are scored based on their similarity to a pre-defined validation set. However, pure attribution methods typically reduce the target task to a single representative signal, effectively favoring samples that align with a centroid of the target representation. This point-wise view fails to capture the geometric structure of the target distribution, leading to selections that are highly redundant and concentrated in a narrow region of the data space. As a result, attribution alone is insufficient for constructing effective training sets.

In response, we propose \method\ in this paper, which synergizes data attribution and diversity into a unified, principled framework for effective data selection in targeted task adaptation. Given a pool of generated candidate samples in each iteration, \method\ assesses task relevance through an optimal-transport-based alignment and promotes diversity by favoring samples that provide complementary training signals. A key component of \method\ is the use of optimal transport (OT)~\citep{peyre2019computational, flamary2021pot}
\footnote{More background on OT can be found in Appendix~\ref{app:addition-prelim}.} 
for target distribution alignment. Instead of matching a single centroid as in attribution-only methods~\citep{xia2024less, kwon2023datainf, san2024in2core}, OT  aligns the training and full target distribution. In Figure~\ref{fig:intro-fig}, we demonstrate that due to the naive selection procedure, attribution-only methods may only select data according to the largest cluster (i.e. the \textcolor{purple}{\textit{red}} cluster) and fail to cover the entire target distribution. 
% \hh{interesting insight. is it possible we can illustrate this difference by a figure?} \tw{sure, will think about this} 
\method\ employs an interleaved optimization procedure that iteratively refines sample importance by jointly optimizing target alignment and diversity, synergizing  two signals in the iterative process.

% \sr{It seems that the main reason for calling our method “EVO” is its iterative nature. Do we have a deeper connection to the idea of evolution in other aspects as well?}\tw{@sirui, do you think this is still a concern? Evo is about setting, not specific to our method}\sr{Yeah, it looks better now. And I would suggest emphasizing more on the "joint nature" of our method as a direct response to the previous paragraph.}

To demonstrate the superiority of \method, we conduct evaluation across 10 knowledge-intensive benchmarks with both weak/strong data generators. Experimental results show that \method\ outperforms strong data selection baselines and even training on all generated data across different selection ratios and iterations. Notably, \method\ is the only approach that consistently improves upon the base model performance throughout the evolution process, preventing harmful adaptation.

%% file: sections/02Prelim.tex
\section{Preliminaries}

\subsection{Training Data Attribution}
\label{ssec:prelim-tda}
% Let $S = \{x_1,\dots,x_n\}$ denote a set of training examples drawn from a space $\mathcal{X}_{\text{train}}$, and let 
% $T = \{x'_1,\dots,x'_m\}$ denote a set of target examples drawn from a space $\mathcal{X}_{\text{validation}}$. 
% We consider a generative model $f_\Theta$ parameterized by $\Theta$ and the data samples here correspond to paired text data (e.g., instructions-response pairs for supervised fine-tuning). Training data attribution aims to quantify how individual training examples in $S$ relate to the model’s behavior on a target example $x' \in T$. Formally, an attribution method $\tau$ assigns a score vector $\tau(x', S; f_\Theta) \in \mathbb{R}^n$,
% where the $i$-th entry $\tau(x', S; f_\Theta)_i$ reflects the degree to which training example $x_i$ is associated with the model’s output on target sample $x'$.

Let $S = \{x_1,\dots,x_n\}$ denote a set of training examples drawn from a space $\mathcal{X}_{\text{train}}$, and let 
$T = \{x'_1,\dots,x'_m\}$ denote a set of target examples drawn from a space $\mathcal{X}_{\text{validation}}$. Each data sample consists of a query–answer pair. We consider a generative model $f_\Theta$ parameterized by $\Theta$. Typically, fine-tuning starts from a pretrained base model with parameters $\Theta_0$, and the adapted parameter $\Theta$ is obtained by minimizing a task-specific objective over the dataset $S$. Training data attribution~\citep{deng2025survey} aims to quantify how individual training examples in $S$ relate to the behavior of $f_\Theta$ on target validation examples in $T$. Formally, an attribution method $\tau$ assigns a score  $\tau(S; f_\Theta) \in \mathbb{R}^n$, where the $i$-th entry $\tau(S; f_\Theta)_i$ reflects the degree to which training example $x_i$ is associated with the model’s output on the target samples. For data selection methods that employ training data attribution, a common strategy is to select the top-$k$ training examples according to their attribution scores. Formally, letting $\tau_i$ denote the average attribution score of training example $x_i$ over $T$, the selected subset can be written as: $S_{\mathrm{sel}}=\underset{S' \subseteq S,\, |S'|=k}{\arg\max}
\sum_{x_i \in S'} \tau(S; f_\Theta)_i$.

\subsection{Problem Definition}

% We study the problem of adapting a base LLM $f_{\Theta_0}$ to a target task distribution $\mathcal{D}_{\text{target}}$ via an \emph{iterative generation--selection--training loop}. 
% We assume access to a small seed set $\mathcal{D}_{\text{seed}}$ from the target task, together with a validation set $\mathcal{D}_{\text{val}}$. 
% At each iteration $i$, the base LLM is provided with a set of candidate examples $\mathcal{D}_{\text{cand}}^{(i)}$ synthesized by an external data generator. For each iteration,a subset $\mathcal{S}_i \subseteq \mathcal{D}_{\text{cand}}^{(i)}$ is \textbf{selected} under a fixed  budget, fine-tune the model using the accumulated dataset (i.e. $\mathcal{D}_{\text{train}}^{(i)} = \mathcal{D}_{\text{seed}} \cup \bigcup_{t=1}^{i} \mathcal{S}_t$) to obtain $f_{\Theta_i}$, and repeat the process. The objective is to optimize the data selection strategy such that the fine-tuned models $\{f_{\Theta_t}\}_{t=1\cdots i}$ achieve maximum accuracy on a hold-out test set from the target task.

We study the problem of adapting a base LLM $f_{\Theta_0}$ to a targeted task distribution $\mathcal{D}_{\text{target}}$ via an \emph{iterative generation--selection--training loop}. 
We assume access to a small seed set $\mathcal{D}_{\text{seed}}$ from the targeted task, together with a validation set $\mathcal{D}_{\text{val}}$. 
At each iteration $t$, the base LLM is provided with a set of candidate examples $\mathcal{D}_{\text{cand}}^{(t)}$ synthesized by an external data generator. 
A subset $\mathcal{S}_t \subseteq \mathcal{D}_{\text{cand}}^{(t)}$ is selected under a fixed budget, and the model is fine-tuned on the accumulated dataset 
(i.e., $\mathcal{D}_{\text{train}}^{(t)} = \mathcal{D}_{\text{seed}} \cup \bigcup_{p=1}^{t} \mathcal{S}_p$) to obtain $f_{\Theta_t}$. 
The process is repeated for $t = 1, \dots, K$. Our objective is to optimize the data selection strategy such that the fine-tuned models $\{f_{\Theta_t}\}_{t=1\cdots K}$ achieve maximum accuracy on a hold-out test set from the targeted task.

%% file: sections/03Method.tex
\section{Methodology}

In this section, we illustrate our method, \method, which selects data to evolve LLM for targeted task adaptation. In Section~\ref{ssec:method-limitation}, we first discuss the limitations of existing methods. Motivated by these limitations, we then introduce \method\ in Section~\ref{ssec:method-detail}. Specifically, \method\ inherently considers task alignment and diversity simultaneously and synergies both signals into a unified, principled selection framework for LLM adaptation.

% \begin{figure}[t]
% \centering
% \includegraphics[width=0.5\linewidth]{COLM 2026/figures/fig1_score_distributions.pdf}
% \caption{win rate by category.}
% \label{fig:my_figure}
% \end{figure}

% \begin{figure}[t]
% \centering
% \includegraphics[width=0.5\linewidth]{COLM 2026/figures/fig3_vendi_bar_ratio02.pdf}
% \caption{win rate by category.}
% \label{fig:my_figure}
% \end{figure}

\subsection{Limitations of Existing Methods}

\label{ssec:method-limitation}

% \paragraph{Problem Setup.}
% We consider the problem of selecting a subset of training samples that is both \emph{target-relevant} and \emph{non-redundant}. 
% Let $\mathcal{S}=\{s_i\}_{i=1}^n$ denote the candidate training set and let $\mathcal{V}=\{v_j\}_{j=1}^m$ denote a validation set that represents the target distribution of interest. Each example is associated with a gradient representation computed with respect to the current model parameters. Specifically, let $\mathbf{g}_i \in \mathbb{R}^d$ denote the gradient feature of training sample $s_i$, and let $\mathbf{h}_j \in \mathbb{R}^d$ denote the gradient feature of validation sample $v_j$. These gradient features encode how each example influences model parameters and therefore provide an attribution signal for measuring task relevance. Rather than selecting samples directly, we maintain a probability weight vector $\mathbf{w} \in \Delta^{n-1}$ over the training samples, where $\Delta^{n-1}=\{\mathbf{w}\in\mathbb{R}_+^n : \sum_i w_i = 1\}$ denotes the probability simplex. The weight $w_i$ indicates the relative importance of sample $s_i$ in representing the target task.

\paragraph{Attribution-Only Selection.}
A popular approach to materialize data selection for targeted task adaptation is through data attribution methods~\citep{deng2025survey, xia2024less, zhang2025survey}, especially gradient-based ones\footnote{According to \citet{deng2025survey}, gradient-based data attribution methods is more favored  over other paradigms for LLM applications due to their efficiency under large scale settings.}. 
% These methods compute gradient similarity~\citep{pruthi2020estimating} or variants of influence functions~\citep{koh2017understanding,kwon2023datainf} to score each training samples w.r.t all validation samples (i.e. 
These methods assign attribution score to each training sample typically via gradient similarity~\citep{pruthi2020estimating} or variants of influence functions~\citep{koh2017understanding, kwon2023datainf, zhou2024hyperinf}. Each training sample is scored by its similarity to validation samples, and the overall score is obtained by averaging over the validation set.

\begin{figure}[t]
\centering

\begin{subfigure}[t]{0.49\linewidth}
    \centering
    \includegraphics[width=\linewidth]{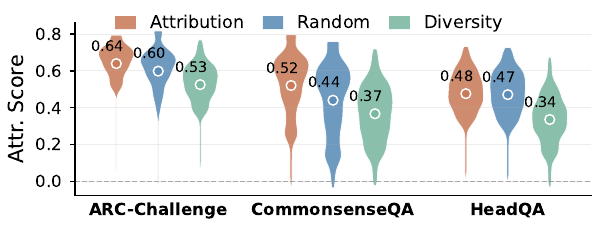}
    \caption{Attribution Score Distribution}
    \label{fig:fig2-attr-score}
\end{subfigure}
\hfill
\begin{subfigure}[t]{0.49\linewidth}
    \centering
    \includegraphics[width=\linewidth]{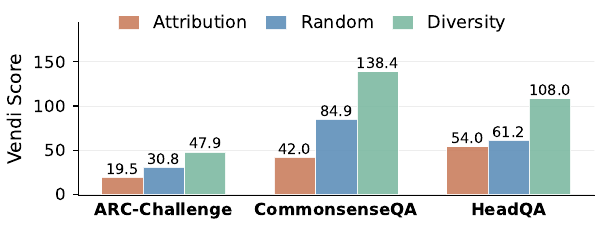}
    \caption{Vendi Score}
    \label{fig:fig2-vendi-score}
\end{subfigure}

\caption{Trade-off between target alignment and diversity. Left: Attribution~\citep{xia2024less} achieves the highest \textit{average attribution score} while Diversity gets the lowest. Right: Diversity~\citep{jung2025prismatic} achieves the highest \textit{Vendi score} while Attribution gets the lowest.}
% \sr{What is vendi score?}\tw{a diversity metric, let me update in the main text}
\label{fig:attr-vendi}
\end{figure}

However, this naive averaging implicitly reduces the validation distribution to a single centroid representation. As a result, samples are ranked based on their proximity to this average signal, without accounting for the geometric structure of the data. When selecting the top-$k$ samples as selected data, this leads to a subset concentrated in a narrow region of the training space. What's worse, this concentration manifests as redundancy: many selected samples exhibit highly similar gradient directions. Consequently, the selected subset provides limited additional coverage of the target distribution, leading to \textit{inefficient use of the selection budget} and even \textit{diluting the effectiveness of training}. We compute the Vendi Score~\citep{friedman2022vendi}, a widely-used diversity metric, to showcase this redundancy issue (the higher, the less redundant). As shown in Figure~\ref{fig:attr-vendi}, while attribution-only selection can successfully identify task-aligned samples (Fig. \ref{fig:fig2-attr-score}), the selected samples exhibit the lowest Vendi Score (i.e. the least diverse)
% \footnote{Vendi Score is a widely-used diversity metric that computes the effective number of unique elements.} ) 
among baselines (Fig. \ref{fig:fig2-vendi-score}).   

% Many existing methods \hh{cite} rank training samples according to their data influence on the validation sample, often measured through , or other attribution metrics. While these methods effectively identify examples that are individually useful for the target task, they often overlook redundancy among selected samples. In practice, multiple training examples may share highly similar gradient directions or represent the same underlying pattern. Selecting all of them leads to a subset that contains redundant information and therefore fails to efficiently utilize the selection budget \hh{then, some might say: let's just select more samples according to relevance (more than the budget), it will still lead to superior performance right?} \tw{yes, the budget is not the point here; will update to argue the issue of training signal dilution (which might harm the model learning)}. As a result, purely attribution-based selection may concentrate heavily on a narrow region of the training distribution.

\paragraph{Diversity-Only Selection.}
On the other hand, diversity-based selection methods~\citep{wu2023self,wang2024diversity,jung2025prismatic} attempt to address redundancy by encouraging coverage of the data space. These approaches typically select samples that maximizes the pairwise embedding distance~\citep{wu2023self} or kernel similarity measure~\citep{wang2024diversity}. While such methods promote coverage of the training distribution, they do not explicitly account for the target task represented by the validation set. Consequently, the selected subset may be diverse but poorly aligned with the target objective, which may potentially degrade performance. In Figure~\ref{fig:attr-vendi}, we can observe that the diversity-only method selects training samples that are the least aligned with the task (Fig. \ref{fig:fig2-attr-score}), even though it achieves the highest Vendi Score (i.e. the most diverse) among all the baselines (Fig. \ref{fig:fig2-vendi-score}).

% In scenarios involving distribution shift or domain adaptation, this limitation becomes especially problematic because covering the entire training distribution does not necessarily help the model generalize to the target domain.

% \tw{figure of attribution score comparing attribution and diversity}

% \tw{figure of vendi score comparing attribution and diversity}

% \sr{Do we compare against methods that jointly consider attribution and diversity, such as TSDS? As written, this section naturally raises the question of how our method relates to prior approaches that already account for both.}\tw{thank you, will look at the TSDS codebase}

\subsection{\method: Synergizing Data Attribution and Diversification}

\label{ssec:method-detail}
% \tw{under development, will update soon}

In this section, we detail our design of \method. To efficiently represent each data sample, we first utilize a proxy model to collect gradient features for both training and validation samples. Then, we apply an OT-based iterative algorithm to simultaneously optimize the targeted task alignment and coverage over the data representation space. 

\paragraph{Proxy Model and Data Representations.}

% things about gradient feature (from LESS) 

% things about proxy model and random projection module (sjlt and dattri)

% Inspired by recent gradient-based data attribution methods~\citep{charpiat2019input,pruthi2020estimating,deng2025survey}, which score each training samples  via gradient similarity and select top-$k$ samples for training. Specifically, they instantiate their attribution function $\tau$ (refer to Section~\ref{ssec:prelim-tda} for details). Let $\ell(x;\Theta)$ denote the loss of model $f_\Theta$ on sample $x$. The attribution of $x_i \in S$ w.r.t. a validation sample $x' \in T$ is approximated by $\tau(x_i, x') \propto \nabla_\Theta \ell(x_i;\Theta)^\top \nabla_\Theta \ell(x';\Theta)$ and thus  $\tau_i = \frac{1}{m} \sum_{x' \in T} \nabla_\Theta \ell(x_i;\Theta)^\top \nabla_\Theta \ell(x';\Theta)$ by averaging over all validation samples. However, computing per-sample gradients requires a full backward pass per example, leading to a computational cost similar to training on full samples.

Recent gradient-based attribution methods~\citep{charpiat2019input,pruthi2020estimating} approximate sample influence via the dot product of gradient features. Formally, the similarity between training sample $x$ and validation sample $x'$ is given by $ \nabla_\Theta \ell(x_i;\Theta)^\top \nabla_\Theta \ell(x';\Theta)$, where $\ell(x;\Theta)$ denotes the loss function. Motivated by this, we represent each data by its gradient feature: $\nabla_\Theta \ell(x;\Theta)$. However, collecting per-sample gradients requires a full backward pass per example, leading to a prohibitive computational cost.

To overcome this limitation, we utilize a proxy model~\citep{engstrom2024dsdm, jung2025prismatic, diao2025nemotron} to collect approximate gradient representations, where the proxy model is much smaller than the base LLM. In addition, the representations are compressed through projection techniques~\footnote{We adopt \textit{Sparse Johnson-Lindenstrauss Transform}~\citep{hu2025grass} in \texttt{dattri}~\citep{deng2024texttt}.} that efficiently preserve pair-wise distances. We also normalize these compressed gradients. Formally, let $\ell(x;\overline\Theta)$ denote the loss under the proxy model. For each sample $x$, its representation is defined as
$\mathbf{g}(x) \;=\; \mathcal{T}\!\left( \nabla_{\overline\Theta} \ell(x;\overline\Theta) \right)$,
where $\mathcal{T}(\cdot)$ is a transformation function that composes projection and normalization. Applying this to the dataset yields training features $\{\mathbf{g}_i^{\mathrm{tr}}\}_{i=1}^n$ and validation features $\{\mathbf{g}_j^{\mathrm{val}}\}_{j=1}^m$.

% and obtain training features $\{\mathbf{g}_i^{\mathrm{tr}}\}_{i=1}^n$ for $n$ training samples and validation features $\{\mathbf{g}_j^{\mathrm{val}}\}_{j=1}^m$ for $m$ validation samples.

% \sr{$\mathbf{g}$ is not defined yet. previously you used $\nabla$?} \tw{modified}

% We then aggregate over the validation set to obtain
% $\tau_i = \frac{1}{|T|} \sum_{x' \in T} \tau(x_i, x') $.
% \sr{It seems like you do not use this score later?}\tw{yeah originally i just want to say how they do this, but actually i only need the motivation of using grad feature}

% Gradient-based data attribution methods~\citep{charpiat2019input,koh2017understanding,pruthi2020estimating} instantiate their $\tau$ (refer to Section~\ref{ssec:prelim-tda} for details) via gradient similarity . Let $\ell(x;\Theta)$ denote the loss of model $f_\Theta$ on sample $x$. The attribution of $x_i \in S$ w.r.t. a validation sample $x' \in T$ is approximated by
% \[
% \tau(x_i, x') \propto \nabla_\Theta \ell(x_i;\Theta)^\top \nabla_\Theta \ell(x';\Theta) \text{ and thus } \tau_i = \frac{1}{m} \sum_{x' \in T} \nabla_\Theta \ell(x_i;\Theta)^\top \nabla_\Theta \ell(x';\Theta).
% \]
% Aggregating over $T$,
% \[
% \tau_i = \frac{1}{m} \sum_{x' \in T} \nabla_\Theta \ell(x_i;\Theta)^\top \nabla_\Theta \ell(x';\Theta),
% \]
% and the selected subset is
% \[
% S_{\mathrm{sel}} = \underset{S' \subseteq S,\, |S'|=k}{\arg\max} \sum_{x_i \in S'} \tau_i.
% \]

\input{tables/main_alg}

\paragraph{Targeted Task Alignment via Optimal Transport.}
% To address these limitations, we first introduce an attribution-driven relevance objective based on optimal transport (OT). Optimal transport provides a principled framework for comparing two probability distributions by computing the minimal cost required to transport probability mass from one distribution to another. 

As discussed in the Introduction section and Section~\ref{ssec:method-limitation}, prior attribution-based methods tend to concentrate selected samples around the centroid of the target task distribution, leading to limited coverage and increased redundancy. To address this limitation, we adopt optimal transport (OT)~\citep{peyre2019computational} to align the synthetic training sample distribution with the validation distribution. Unlike centroid matching, OT respects the full geometry of the distributions, enabling a more comprehensive and globally consistent coverage of the target task space. 

Formally, let $\mathbf{w} \in \Delta^{n-1}$ denote the weight distribution over training samples and let $\mathbf{b} = \frac{1}{m}\mathbf{1}_m \in \Delta^{m-1}$ denote the uniform distribution over validation samples. 
We define the cost matrix $\mathbf{C}=[C_{ij}]=\|\mathbf{g}_i^{\mathrm{tr}}-\mathbf{g}_j^{\mathrm{val}}\|_2^2$ that measures the pairwise distance between training and validation features\footnote{With normalized data representations, $\|\mathbf{g}_i^{\mathrm{tr}}-\mathbf{g}_j^{\mathrm{val}}\|_2^2 = 2 - 2\langle \mathbf{g}_i^{\mathrm{tr}}, \mathbf{g}_j^{\mathrm{val}} \rangle$, so our cost matrix is equivalent (up to scaling and shift) to the one induced by the inner product.} 
% \hh{so, we use the gradient info as the representation. what is the intuition/rationale of using such info to define the cost matrix?}\tw{thanks for the notice, i should mention the Grad-Sim papers in the first paragraph; after we know the grad-similarity is a useful metric, the cost matrix definition should be straight-forward as shown in the footnote}. 
The optimal transport distance between the two distributions is then defined as $\mathrm{OT}(\mathbf{w},\mathbf{b}) =
\min_{\mathbf{P}\in\Pi(\mathbf{w},\mathbf{b})}
\sum_{i=1}^{n}\sum_{j=1}^{m} P_{ij} C_{ij}$,
where $\mathbf{P}$ is a transport plan and $\Pi(\mathbf{w},\mathbf{b})$ denotes the set of valid couplings\footnote{We adopt entropic-regularized OT, which can be solved efficiently via the Sinkhorn algorithm~\citep{sinkhorn1967diagonal}. More background on OT is in Appendix~\ref{app:addition-prelim}.}. 
% Intuitively, minimizing the OT distance encourages selecting training samples whose gradients align with those of the validation data, which represents the target task distribution. 
% Importantly, the dual potential $\mathbf{u}$ \hh{what is u?}\tw{it is the dual solution of the weight vector $\textbf{w}$} of the primal OT problem provides the gradient of the OT objective with respect to training weight $\mathbf{w}$. 
Importantly, we can efficiently compute the gradient of the OT objective with respect to the training weights $\mathbf{w}$ by leveraging its dual formulation. Specifically, let $\mathbf{u}$
% \hh{what is u?}\tw{I update the entire sentence}  
be the optimal dual variable associated with the training marginal constraint $\mathbf{w}$. Following the sensitivity theorem~\citep{bertsekas1997nonlinear}, we can obtain: $\nabla_{\mathbf{w}} \mathrm{OT}(\mathbf{w},\mathbf{b}) = \mathbf{u}$.

% Firstly, let $\mathbf{w}$ denote the weight distribution over training samples and let $\mathbf{b} = \frac{1}{m}\mathbf{1}_{m}\in\Delta^{m-1}$ denote the (uniform) validation distribution. We define the cost matrix

% The dual potentials of this problem yield vectors $\mathbf{u}\in\mathbb{R}^n$ and $\mathbf{v}\in\mathbb{R}^m$ satisfying
% \[
% \mathrm{OT}_\varepsilon(\mathbf{w},\mathbf{b})
% =
% \mathbf{u}^{\top}\mathbf{w}
% +
% \mathbf{v}^{\top}\mathbf{b}.
% \]

\paragraph{Diversity Regularization.}
To mitigate the accumulated redundancy of generated training candidates  throughout the iterative procedure, we introduce a diversity regularization term that penalizes selecting samples with highly similar gradients. Let $\mathbf{S} = [S_{ii'}] = \langle \mathbf{g}^{\text{tr}}_i, \mathbf{g}^{\text{tr}}_{i'} \rangle$
denotes a similarity kernel between training samples. We define the diversity energy: $E_{\text{div}}(\mathbf{w})
=
\frac{1}{2}\mathbf{w}^{\top} S \mathbf{w}$,
whose gradient is given by $\nabla_\textbf{w} E_{\text{div}}(\mathbf{w}) = S\mathbf{w}$.
This gradient measures how crowded the neighborhood of each sample is in gradient space. Thus, if the OT objective assigns high-weight to a sample's similar neighbors in the representation space, the diversity gradient will discourage further increases in its own weight.

\paragraph{Unified Optimization.}
We summarize the entire procedure of \method\ in Algorithm~\ref{alg:method-algorithm-block}. 
% In brief, \method\ integrates the distributional alignment signal and the diversity regularization in order to select a task-aligned, diverse subset for effective LLM adaption. 
% \hh{refer to key step(s) in Algorithm 1} \tw{added}
At each iteration $t$ within the \textit{generation-selection-training loop}, we compute the OT gradient $\mathbf{u}^{(t)}$ together with the diversity gradient $\mathbf{d}^{(t)}$, and combine them to form a joint gradient $\mathbf{g}^{(t)}$ (\textit{line 6-8)}. We then update the weight vector $\mathbf{w}^{(t)}$ for the synthetic samples generated in iteration $t$ with an exponentiated update, followed by a normalization to stay on the probability simplex (\textit{line 9-10)}. After $T$ runs of optimization, the final subset is obtained by selecting the top-$k$ samples with the largest weights, where $k = \lfloor n\rho \rfloor$ corresponds to the desired  budget under selection ratio $\rho$ (\textit{line 12-13)}. 

\paragraph{Complexity Analysis of \method.}

Let $n$ be the number of training samples to be selected in the current iteration, $m$ be the number of validation samples, $\gamma$ be the approximation error introduced by approximate Sinkhorn solvers~\citep{sinkhorn1967diagonal,dvurechensky2018computational}, and $T$ be the number of update steps in \method. The runtime complexity can be summarized in the following proposition. 
% The empirical runtime can be found in Appendix xxx. \tw{add runtime?}

\begin{proposition}[Time Complexity Analysis~\citep{dvurechensky2018computational}]
\method\ has the time complexity $\mathcal{O}(Tnm \log (\max (n,m) ) \gamma^{-2})$.

\end{proposition}

%% file: tables/main_alg.tex
\begin{algorithm}[t]
\caption{\method}
\label{alg:method-algorithm-block}
\begin{algorithmic}[1]
\Require Training set $\mathcal{S}=\{s_i\}_{i=1}^n$, training features $\{\mathbf{g}_i^{\mathrm{tr}}\}_{i=1}^n$, validation features $\{\mathbf{g}_j^{\mathrm{val}}\}_{j=1}^m$, selection ratio $\rho$, number of steps $T$, learning rate $\eta$, regularization coefficient $\varepsilon$
\Ensure Selected subset $\mathcal{S}_{\mathrm{sel}}$
\State $k \gets \max(1,\lfloor n\rho \rfloor)$
\State Construct cost matrix $\mathbf{C}=[C_{ij}]=\|\mathbf{g}_i^{\mathrm{tr}}-\mathbf{g}_j^{\mathrm{val}}\|_2^2 \in \mathbb{R}^{n\times m}$
\State Construct similarity matrix $\mathbf{S}=[S_{ii'}]=\langle \mathbf{g}_i^{\mathrm{tr}},\mathbf{g}_{i'}^{\mathrm{tr}}\rangle\in \mathbb{R}^{n\times n}$
\State Initialize validation marginal $\mathbf{b}\gets \frac{1}{m}\mathbf{1}_{m}$ and training weights $\mathbf{w}^{(0)}\gets \frac{1}{n}\mathbf{1}_{n}$
\For{$t=0,\dots,T-1$}
    % \State $\mathbf{p}^{(t)} \gets \mathbf{w}^{(t)} / \|\mathbf{w}^{(t)}\|_1$
    \State $\mathbf{u}^{(t)} \gets \nabla_{\mathbf{w}^{(t)}}\mathrm{OT}(\mathbf{w}^{(t)},\mathbf{b},\mathbf{C},\varepsilon)$ \Comment{OT gradient}
    \State $\mathbf{d}^{(t)} \gets \mathbf{S}\mathbf{w}^{(t)}$ \Comment{diversity gradient}
    \State $\mathbf{g}^{(t)} \gets \mathrm{Standardize}(\mathbf{u}^{(t)}) + \mathrm{Standardize}(\mathbf{d}^{(t)})$
    \State $\mathbf{w}^{(t+1)} \gets \mathbf{w}^{(t)} \odot \exp(-\eta \mathbf{g}^{(t)})$ \Comment{exponentiated update}
    \State $\mathbf{w}^{(t+1)} \gets \mathbf{w}^{(t+1)} / \|\mathbf{w}^{(t+1)}\|_1$ \Comment{re-normalize training weights}
\EndFor
\State Let $\mathcal{I}_{\mathrm{top}}$ be the indices of the top-$k$ entries of $\mathbf{w}^{(T)}$
\State \Return $\mathcal{S}_{\mathrm{sel}} \gets \{s_i : i \in \mathcal{I}_{\mathrm{top}}\}$
\end{algorithmic}
\end{algorithm}

%% file: sections/04Experiments.tex
\section{Experiment}
\label{sec:exp}

\subsection{Experimental Setup}

\paragraph{General setting.}

We consider the \textit{generation–selection–training} setting. At each iteration, an external data generator produces candidate examples based on the current seed set and a fixed validation set. A selection method is then applied to construct a training subset under a specified selection ratio, after which the base LLM is fine-tuned on the selected data. In this paper, we run this process for \textit{two} iterations. The construction for seed/validation sets and partition for each dataset is detailed in Appendix~\ref{app:dataset-settings}. 
% At each iteration, 5k problem instance is generated and for each instance, three candidate rationales and solutions are generated~\citep{zelikman2022star}.
% The candidate solutions will then be passed through a majority voting filter. 
The problem/answer generation configurations and prompts are included in Appendix~\ref{app:gen-prompts}. The detailed training configuration is in Appendix~\ref{app:training-config}.

% At each iteration, the system generates candidate problems. For each problem, three answers are generated and filtered using majority voting. The selected training set for the iteration contains up to 5{,}000 examples. 

\paragraph{Datasets.}
We consider a diverse set of benchmarks spanning science, commonsense, logical  and biomedical question answering datasets. These datasets include (i) scientific reasoning: \texttt{ARC-Challenge}~\citep{clark2018think}, \texttt{MMLU}~\citep{hendrycks2020measuring},
\texttt{OpenBookQA}~\citep{mihaylov2018can} and  \texttt{ClimaQA}~\citep{manivannan2024climaqa}; (ii) commonsense/logical reasoning: \texttt{CommonsenseQA}~\citep{talmor2019commonsenseqa}, \texttt{LogiQA}~\citep{liu2020logiqa} and \texttt{LogiQA2}~\citep{liu2023logiqa}; (iii) biomedical/healthcare reasoning: \texttt{Med-MCQA}~\citep{pal2022medmcqa}, \texttt{MedQA}~\citep{jin2021disease} and \texttt{HeadQA}~\citep{vilares2019head}. The prompts and detailed configuration for evaluation is in Appendix~\ref{app:evaluation}.

\paragraph{Models.}
We primarily conduct experiments with the Qwen2.5 family~\citep{qwen2.5} due to its various model sizes and wild adoption. The data generator is fixed to \texttt{Qwen2.5-14B-Instruct}, which is used to generate questions and solutions. The LLM to be trained (i.e. the base model) is either \texttt{Qwen2.5-3B-Instruct} (i.e. strong generator setting) or \texttt{Qwen2.5-14B-Instruct} (i.e. weak generator setting), and we use \texttt{Qwen2.5-0.5B-Instruct} as the proxy model to collect gradient features.

% All evaluations are conducted in a zero-shot setting with greedy decoding (temperature $0.0$).

\paragraph{\method\ and baseline selection methods.}
For \method, we fix learning rate $\eta = 0.1$, consider update steps $T \in \{10, 20\}$ and regularization coefficient $\epsilon \in \{0.5,1.0\}$. The selection ratios are $0.2$ and $0.5$. We also evaluate several baseline methods. First, we include simple methods: \textit{All} and \textit{Random}, which train the base model on full and random subsets. Second, we consider \textit{Diversity}, which selects examples by prioritizing smaller clusters~\citep{jung2025prismatic}, and \textit{Attribution}~\citep{xia2024less}, which favors samples with high gradient similarity scores against validation data. For methods that incorporate diversity and task relevance, we consider \textit{TSDS}~\citep{liu2024tsds} as one representative method. We also compare against a baseline, \textit{Attr-Div}, that combines attribution with diversity in a sequential manner. We describe further details about these baseline methods in Appendix~\ref{app:baselines}.

\begin{table*}[t]
\centering
\renewcommand{\arraystretch}{0.75}
\resizebox{1.0\textwidth}{!}{%
\begin{tabular}{c l cc cc cc cc cc}
\toprule
\midrule
 &  & \multicolumn{2}{c}{\textbf{ARC-Challenge}} & \multicolumn{2}{c}{\textbf{MMLU}} & \multicolumn{2}{c}{\textbf{OpenBookQA}} & \multicolumn{2}{c}{\textbf{ClimaQA}} & \multicolumn{2}{c}{\textbf{\textbf{Avg.}}} \\
\cmidrule(lr){3-4} \cmidrule(lr){5-6} \cmidrule(lr){7-8} \cmidrule(lr){9-10} \cmidrule(lr){11-12}
 & \textbf{Selection Ratio} $\rightarrow$ & \multicolumn{1}{c}{0.2} & \multicolumn{1}{c}{0.5} & \multicolumn{1}{c}{0.2} & \multicolumn{1}{c}{0.5} & \multicolumn{1}{c}{0.2} & \multicolumn{1}{c}{0.5} & \multicolumn{1}{c}{0.2} & \multicolumn{1}{c}{0.5} & \multicolumn{1}{c}{0.2} & \multicolumn{1}{c}{0.5} \\
\cmidrule{2-12}
 & \textbf{Methods} $\downarrow$ & & & & & & & & & & \\
\cmidrule{2-12}
 & \cellcolor{BaseRow}Base & \multicolumn{2}{c}{\cellcolor{BaseRow}0.8140} & \multicolumn{2}{c}{\cellcolor{BaseRow}0.6548} & \multicolumn{2}{c}{\cellcolor{BaseRow}0.8000} & \multicolumn{2}{c}{\cellcolor{BaseRow}0.7329} & \multicolumn{2}{c}{\cellcolor{BaseRow}0.7504} \\
\midrule
\multirow{7}{*}{\textbf{Iter.\ 1}} & \cellcolor{AllRow}All & \multicolumn{2}{c}{\cellcolor{AllRow}0.8055} & \multicolumn{2}{c}{\cellcolor{AllRow}0.6522} & \multicolumn{2}{c}{\cellcolor{AllRow}0.7760} & \multicolumn{2}{c}{\cellcolor{AllRow}0.7329} & \multicolumn{2}{c}{\cellcolor{AllRow}0.7416} \\
 & \cellcolor{BaselineRow}Random & \cellcolor{BaselineRow}\underline{0.8234} & \cellcolor{BaselineRow}0.8063 & \cellcolor{BaselineRow}0.6557 & \cellcolor{BaselineRow}0.6470 & \cellcolor{BaselineRow}0.7780 & \cellcolor{BaselineRow}\textbf{0.8060} & \cellcolor{BaselineRow}\textbf{0.7466} & \cellcolor{BaselineRow}0.7260 & \cellcolor{BaselineRow}\underline{0.7509} & \cellcolor{BaselineRow}0.7463 \\
 & \cellcolor{BaselineRow}Attribution & \cellcolor{BaselineRow}0.8140 & \cellcolor{BaselineRow}0.8080 & \cellcolor{BaselineRow}0.6503 & \cellcolor{BaselineRow}0.6515 & \cellcolor{BaselineRow}0.7960 & \cellcolor{BaselineRow}0.7920 & \cellcolor{BaselineRow}0.7089 & \cellcolor{BaselineRow}\underline{0.7500} & \cellcolor{BaselineRow}0.7423 & \cellcolor{BaselineRow}\underline{0.7504} \\
 & \cellcolor{BaselineRow}Diversity & \cellcolor{BaselineRow}0.8174 & \cellcolor{BaselineRow}\underline{0.8200} & \cellcolor{BaselineRow}\textbf{0.6605} & \cellcolor{BaselineRow}\underline{0.6534} & \cellcolor{BaselineRow}0.7860 & \cellcolor{BaselineRow}0.7860 & \cellcolor{BaselineRow}0.7055 & \cellcolor{BaselineRow}0.7397 & \cellcolor{BaselineRow}0.7424 & \cellcolor{BaselineRow}0.7498 \\
 & \cellcolor{BaselineRow}Attr-Div & \cellcolor{BaselineRow}0.8089 & \cellcolor{BaselineRow}0.8123 & \cellcolor{BaselineRow}0.6457 & \cellcolor{BaselineRow}0.6474 & \cellcolor{BaselineRow}0.7980 & \cellcolor{BaselineRow}\underline{0.7980} & \cellcolor{BaselineRow}0.7021 & \cellcolor{BaselineRow}0.7021 & \cellcolor{BaselineRow}0.7387 & \cellcolor{BaselineRow}0.7399 \\
 & \cellcolor{BaselineRow}TSDS & \cellcolor{BaselineRow}0.8063 & \cellcolor{BaselineRow}0.8123 & \cellcolor{BaselineRow}\underline{0.6562} & \cellcolor{BaselineRow}0.6474 & \cellcolor{BaselineRow}\textbf{0.8200} & \cellcolor{BaselineRow}0.7880 & \cellcolor{BaselineRow}0.7158 & \cellcolor{BaselineRow}0.7295 & \cellcolor{BaselineRow}0.7495 & \cellcolor{BaselineRow}0.7443 \\
 & \cellcolor{OursRow}\textbf{\method} & \cellcolor{OursRow}\textbf{0.8285} & \cellcolor{OursRow}\textbf{0.8208} & \cellcolor{OursRow}0.6539 & \cellcolor{OursRow}\textbf{0.6561} & \cellcolor{OursRow}\underline{0.8100} & \cellcolor{OursRow}\textbf{0.8060} & \cellcolor{OursRow}\underline{0.7397} & \cellcolor{OursRow}\textbf{0.7568} & \cellcolor{OursRow}\textbf{0.7580} & \cellcolor{OursRow}\textbf{0.7599} \\
\midrule
\multirow{7}{*}{\textbf{Iter.\ 2}} & \cellcolor{AllRow}All & \multicolumn{2}{c}{\cellcolor{AllRow}0.8200} & \multicolumn{2}{c}{\cellcolor{AllRow}0.6436} & \multicolumn{2}{c}{\cellcolor{AllRow}0.8040} & \multicolumn{2}{c}{\cellcolor{AllRow}0.7466} & \multicolumn{2}{c}{\cellcolor{AllRow}0.7535} \\
 & \cellcolor{BaselineRow}Random & \cellcolor{BaselineRow}0.8157 & \cellcolor{BaselineRow}0.8183 & \cellcolor{BaselineRow}0.6498 & \cellcolor{BaselineRow}0.6486 & \cellcolor{BaselineRow}0.7980 & \cellcolor{BaselineRow}\underline{0.8020} & \cellcolor{BaselineRow}0.7158 & \cellcolor{BaselineRow}0.7055 & \cellcolor{BaselineRow}0.7448 & \cellcolor{BaselineRow}0.7436 \\
 & \cellcolor{BaselineRow}Attribution & \cellcolor{BaselineRow}0.8183 & \cellcolor{BaselineRow}0.8234 & \cellcolor{BaselineRow}0.6495 & \cellcolor{BaselineRow}0.6508 & \cellcolor{BaselineRow}0.7800 & \cellcolor{BaselineRow}\underline{0.8020} & \cellcolor{BaselineRow}0.6986 & \cellcolor{BaselineRow}\textbf{0.7534} & \cellcolor{BaselineRow}0.7366 & \cellcolor{BaselineRow}\underline{0.7574} \\
 & \cellcolor{BaselineRow}Diversity & \cellcolor{BaselineRow}0.8191 & \cellcolor{BaselineRow}\underline{0.8268} & \cellcolor{BaselineRow}\underline{0.6528} & \cellcolor{BaselineRow}0.6492 & \cellcolor{BaselineRow}0.7660 & \cellcolor{BaselineRow}0.7980 & \cellcolor{BaselineRow}0.7123 & \cellcolor{BaselineRow}0.7466 & \cellcolor{BaselineRow}0.7376 & \cellcolor{BaselineRow}0.7551 \\
 & \cellcolor{BaselineRow}Attr-Div & \cellcolor{BaselineRow}0.8106 & \cellcolor{BaselineRow}0.8131 & \cellcolor{BaselineRow}0.6488 & \cellcolor{BaselineRow}\underline{0.6542} & \cellcolor{BaselineRow}0.7880 & \cellcolor{BaselineRow}0.7940 & \cellcolor{BaselineRow}0.7123 & \cellcolor{BaselineRow}0.7329 & \cellcolor{BaselineRow}0.7399 & \cellcolor{BaselineRow}0.7486 \\
 & \cellcolor{BaselineRow}TSDS & \cellcolor{BaselineRow}\underline{0.8229} & \cellcolor{BaselineRow}\textbf{0.8345} & \cellcolor{BaselineRow}0.6478 & \cellcolor{BaselineRow}0.6492 & \cellcolor{BaselineRow}\underline{0.8060} & \cellcolor{BaselineRow}0.7920 & \cellcolor{BaselineRow}0.7158 & \cellcolor{BaselineRow}0.7260 & \cellcolor{BaselineRow}\underline{0.7506} & \cellcolor{BaselineRow}0.7504 \\
 & \cellcolor{OursRow}\textbf{\method} & \cellcolor{OursRow}\textbf{0.8242} & \cellcolor{OursRow}0.8191 & \cellcolor{OursRow}\textbf{0.6586} & \cellcolor{OursRow}\textbf{0.6552} & \cellcolor{OursRow}\textbf{0.8080} & \cellcolor{OursRow}\textbf{0.8080} & \cellcolor{OursRow}\textbf{0.7432} & \cellcolor{OursRow}\underline{0.7500} & \cellcolor{OursRow}\textbf{0.7585} & \cellcolor{OursRow}\textbf{0.7581} \\
\midrule
\bottomrule
\end{tabular}
}%
\caption{Performance comparison on scientific datasets (\texttt{3b} base model). \textbf{Bold} \& \underline{underscore} denote top-1/2 accuracy.}
\label{tab:strong-table-1}
\end{table*}

\begin{table*}[!t]
\centering
\renewcommand{\arraystretch}{0.75}
\resizebox{1.0\textwidth}{!}{%
\begin{tabular}{c l cc cc cc cc}
\toprule
\midrule
 &  & \multicolumn{2}{c}{\textbf{CommonsenseQA}} & \multicolumn{2}{c}{\textbf{LogiQA}} & \multicolumn{2}{c}{\textbf{LogiQA2}} & \multicolumn{2}{c}{\textbf{\textbf{Avg.}}} \\
\cmidrule(lr){3-4} \cmidrule(lr){5-6} \cmidrule(lr){7-8} \cmidrule(lr){9-10}
 & \textbf{Selection Ratio} $\rightarrow$ & \multicolumn{1}{c}{0.2} & \multicolumn{1}{c}{0.5} & \multicolumn{1}{c}{0.2} & \multicolumn{1}{c}{0.5} & \multicolumn{1}{c}{0.2} & \multicolumn{1}{c}{0.5} & \multicolumn{1}{c}{0.2} & \multicolumn{1}{c}{0.5} \\
\cmidrule{2-10}
 & \textbf{Methods} $\downarrow$ & & & & & & & & \\
\cmidrule{2-10}
 & \cellcolor{BaseRow}Base & \multicolumn{2}{c}{\cellcolor{BaseRow}0.7510} & \multicolumn{2}{c}{\cellcolor{BaseRow}0.4670} & \multicolumn{2}{c}{\cellcolor{BaseRow}0.3836} & \multicolumn{2}{c}{\cellcolor{BaseRow}0.5339} \\
\midrule
\multirow{7}{*}{\textbf{Iter.\ 1}} & \cellcolor{AllRow}All & \multicolumn{2}{c}{\cellcolor{AllRow}0.7527} & \multicolumn{2}{c}{\cellcolor{AllRow}0.4455} & \multicolumn{2}{c}{\cellcolor{AllRow}0.3995} & \multicolumn{2}{c}{\cellcolor{AllRow}0.5326} \\
 & \cellcolor{BaselineRow}Random & \cellcolor{BaselineRow}0.7666 & \cellcolor{BaselineRow}0.7592 & \cellcolor{BaselineRow}\underline{0.4762} & \cellcolor{BaselineRow}0.4378 & \cellcolor{BaselineRow}\underline{0.4122} & \cellcolor{BaselineRow}0.4084 & \cellcolor{BaselineRow}\underline{0.5517} & \cellcolor{BaselineRow}0.5351 \\
 & \cellcolor{BaselineRow}Attribution & \cellcolor{BaselineRow}0.7690 & \cellcolor{BaselineRow}0.7576 & \cellcolor{BaselineRow}0.4531 & \cellcolor{BaselineRow}0.4562 & \cellcolor{BaselineRow}0.3899 & \cellcolor{BaselineRow}\textbf{0.4179} & \cellcolor{BaselineRow}0.5373 & \cellcolor{BaselineRow}0.5439 \\
 & \cellcolor{BaselineRow}Diversity & \cellcolor{BaselineRow}0.7649 & \cellcolor{BaselineRow}0.7527 & \cellcolor{BaselineRow}0.4747 & \cellcolor{BaselineRow}0.4762 & \cellcolor{BaselineRow}0.3938 & \cellcolor{BaselineRow}0.4001 & \cellcolor{BaselineRow}0.5445 & \cellcolor{BaselineRow}0.5430 \\
 & \cellcolor{BaselineRow}Attr-Div & \cellcolor{BaselineRow}\underline{0.7723} & \cellcolor{BaselineRow}\textbf{0.7764} & \cellcolor{BaselineRow}0.4485 & \cellcolor{BaselineRow}0.4516 & \cellcolor{BaselineRow}0.3906 & \cellcolor{BaselineRow}0.3861 & \cellcolor{BaselineRow}0.5371 & \cellcolor{BaselineRow}0.5380 \\
  & \cellcolor{BaselineRow}TSDS & \cellcolor{BaselineRow}0.7707 & \cellcolor{BaselineRow}0.7527 & \cellcolor{BaselineRow}0.4608 & \cellcolor{BaselineRow}\underline{0.4839} & \cellcolor{BaselineRow}0.4027 & \cellcolor{BaselineRow}0.4020 & \cellcolor{BaselineRow}0.5447 & \cellcolor{BaselineRow}\underline{0.5462} \\
 & \cellcolor{OursRow}\textbf{\method} & \cellcolor{OursRow}\textbf{0.7756} & \cellcolor{OursRow}\underline{0.7690} & \cellcolor{OursRow}\textbf{0.4793} & \cellcolor{OursRow}\textbf{0.4962} & \cellcolor{OursRow}\textbf{0.4148} & \cellcolor{OursRow}\underline{0.4109} & \cellcolor{OursRow}\textbf{0.5566} & \cellcolor{OursRow}\textbf{0.5587} \\
\midrule
\multirow{7}{*}{\textbf{Iter.\ 2}} & \cellcolor{AllRow}All & \multicolumn{2}{c}{\cellcolor{AllRow}0.7420} & \multicolumn{2}{c}{\cellcolor{AllRow}0.4731} & \multicolumn{2}{c}{\cellcolor{AllRow}0.3969} & \multicolumn{2}{c}{\cellcolor{AllRow}0.5373} \\
 & \cellcolor{BaselineRow}Random & \cellcolor{BaselineRow}0.7633 & \cellcolor{BaselineRow}\underline{0.7641} & \cellcolor{BaselineRow}\textbf{0.4977} & \cellcolor{BaselineRow}0.4455 & \cellcolor{BaselineRow}\underline{0.4109} & \cellcolor{BaselineRow}0.3995 & \cellcolor{BaselineRow}\textbf{0.5573} & \cellcolor{BaselineRow}0.5364 \\
 & \cellcolor{BaselineRow}Attribution & \cellcolor{BaselineRow}0.7584 & \cellcolor{BaselineRow}0.7600 & \cellcolor{BaselineRow}0.4547 & \cellcolor{BaselineRow}\textbf{0.4916} & \cellcolor{BaselineRow}0.3766 & \cellcolor{BaselineRow}\underline{0.4103} & \cellcolor{BaselineRow}0.5299 & \cellcolor{BaselineRow}\underline{0.5540} \\
 & \cellcolor{BaselineRow}Diversity & \cellcolor{BaselineRow}0.7609 & \cellcolor{BaselineRow}0.7518 & \cellcolor{BaselineRow}0.4516 & \cellcolor{BaselineRow}0.4639 & \cellcolor{BaselineRow}0.4065 & \cellcolor{BaselineRow}0.4084 & \cellcolor{BaselineRow}0.5397 & \cellcolor{BaselineRow}0.5414 \\
 & \cellcolor{BaselineRow}Attr-Div & \cellcolor{BaselineRow}\underline{0.7764} & \cellcolor{BaselineRow}0.7584 & \cellcolor{BaselineRow}0.4624 & \cellcolor{BaselineRow}0.4624 & \cellcolor{BaselineRow}0.4001 & \cellcolor{BaselineRow}0.3842 & \cellcolor{BaselineRow}0.5463 & \cellcolor{BaselineRow}0.5350 \\
  & \cellcolor{BaselineRow}TSDS& \cellcolor{BaselineRow}0.7559 & \cellcolor{BaselineRow}0.7494 & \cellcolor{BaselineRow}\underline{0.4808} & \cellcolor{BaselineRow}0.4547 & \cellcolor{BaselineRow}0.3989 & \cellcolor{BaselineRow}0.3995 & \cellcolor{BaselineRow}0.5452 & \cellcolor{BaselineRow}0.5345 \\
 & \cellcolor{OursRow}\textbf{\method} & \cellcolor{OursRow}\textbf{0.7805} & \cellcolor{OursRow}\textbf{0.7707} & \cellcolor{OursRow}0.4608 & \cellcolor{OursRow}\underline{0.4854} & \cellcolor{OursRow}\textbf{0.4198} & \cellcolor{OursRow}\textbf{0.4116} & \cellcolor{OursRow}\underline{0.5537} & \cellcolor{OursRow}\textbf{0.5559} \\
\midrule
\bottomrule
\end{tabular}
}%
\caption{Performance comparison on commonsense/logical datasets (\texttt{3b} base model). \textbf{Bold} and \underline{underscore} denote top-1/2 accuracy.}
\label{tab:strong-table-2}
\end{table*}

\begin{table*}[!htbp]
\centering
\renewcommand{\arraystretch}{0.75}
\resizebox{1.0\textwidth}{!}{%
\begin{tabular}{c l cc cc cc cc}
\toprule
\midrule
 &  & \multicolumn{2}{c}{\textbf{Med-MCQA}} & \multicolumn{2}{c}{\textbf{MedQA}} & \multicolumn{2}{c}{\textbf{HeadQA}} & \multicolumn{2}{c}{\textbf{\textbf{Avg.}}} \\
\cmidrule(lr){3-4} \cmidrule(lr){5-6} \cmidrule(lr){7-8} \cmidrule(lr){9-10}
 & \textbf{Selection Ratio} $\rightarrow$ & \multicolumn{1}{c}{0.2} & \multicolumn{1}{c}{0.5} & \multicolumn{1}{c}{0.2} & \multicolumn{1}{c}{0.5} & \multicolumn{1}{c}{0.2} & \multicolumn{1}{c}{0.5} & \multicolumn{1}{c}{0.2} & \multicolumn{1}{c}{0.5} \\
\cmidrule{2-10}
 & \textbf{Methods} $\downarrow$ & & & & & & & & \\
\cmidrule{2-10}
 & \cellcolor{BaseRow}Base & \multicolumn{2}{c}{\cellcolor{BaseRow}0.4731} & \multicolumn{2}{c}{\cellcolor{BaseRow}0.5043} & \multicolumn{2}{c}{\cellcolor{BaseRow}0.6510} & \multicolumn{2}{c}{\cellcolor{BaseRow}0.5428} \\
\midrule
\multirow{7}{*}{\textbf{Iter.\ 1}} & \cellcolor{AllRow}All & \multicolumn{2}{c}{\cellcolor{AllRow}0.4867} & \multicolumn{2}{c}{\cellcolor{AllRow}0.4918} & \multicolumn{2}{c}{\cellcolor{AllRow}0.6477} & \multicolumn{2}{c}{\cellcolor{AllRow}0.5421} \\
 & \cellcolor{BaselineRow}Random & \cellcolor{BaselineRow}\underline{0.4891} & \cellcolor{BaselineRow}0.4781 & \cellcolor{BaselineRow}\underline{0.5263} & \cellcolor{BaselineRow}\underline{0.5208} & \cellcolor{BaselineRow}0.6597 & \cellcolor{BaselineRow}\underline{0.6579} & \cellcolor{BaselineRow}\textbf{0.5584} & \cellcolor{BaselineRow}\underline{0.5523} \\
 & \cellcolor{BaselineRow}Attribution & \cellcolor{BaselineRow}0.4808 & \cellcolor{BaselineRow}0.4817 & \cellcolor{BaselineRow}0.5232 & \cellcolor{BaselineRow}0.5067 & \cellcolor{BaselineRow}0.6608 & \cellcolor{BaselineRow}0.6532 & \cellcolor{BaselineRow}0.5549 & \cellcolor{BaselineRow}0.5472 \\
 & \cellcolor{BaselineRow}Diversity & \cellcolor{BaselineRow}0.4848 & \cellcolor{BaselineRow}0.4769 & \cellcolor{BaselineRow}0.5035 & \cellcolor{BaselineRow}0.5185 & \cellcolor{BaselineRow}0.6524 & \cellcolor{BaselineRow}0.6503 & \cellcolor{BaselineRow}0.5469 & \cellcolor{BaselineRow}0.5486 \\
 & \cellcolor{BaselineRow}Attr-Div & \cellcolor{BaselineRow}0.4772 & \cellcolor{BaselineRow}0.4829 & \cellcolor{BaselineRow}\textbf{0.5302} & \cellcolor{BaselineRow}0.5106 & \cellcolor{BaselineRow}\textbf{0.6659} & \cellcolor{BaselineRow}0.6557 & \cellcolor{BaselineRow}\underline{0.5578} & \cellcolor{BaselineRow}0.5497 \\
  & \cellcolor{BaselineRow}TSDS & \cellcolor{BaselineRow}0.4808 & \cellcolor{BaselineRow}\underline{0.4898} & \cellcolor{BaselineRow}0.5137 & \cellcolor{BaselineRow}0.5043 & \cellcolor{BaselineRow}0.6601 & \cellcolor{BaselineRow}0.6568 & \cellcolor{BaselineRow}0.5515 & \cellcolor{BaselineRow}0.5503 \\
 & \cellcolor{OursRow}\textbf{\method} & \cellcolor{OursRow}\textbf{0.4915} & \cellcolor{OursRow}\textbf{0.4927} & \cellcolor{OursRow}0.5169 & \cellcolor{OursRow}\textbf{0.5334} & \cellcolor{OursRow}\underline{0.6619} & \cellcolor{OursRow}\textbf{0.6590} & \cellcolor{OursRow}0.5568 & \cellcolor{OursRow}\textbf{0.5617} \\
\midrule
\multirow{7}{*}{\textbf{Iter.\ 2}} & \cellcolor{AllRow}All & \multicolumn{2}{c}{\cellcolor{AllRow}0.4793} & \multicolumn{2}{c}{\cellcolor{AllRow}0.5035} & \multicolumn{2}{c}{\cellcolor{AllRow}0.6652} & \multicolumn{2}{c}{\cellcolor{AllRow}0.5493} \\
 & \cellcolor{BaselineRow}Random & \cellcolor{BaselineRow}0.4843 & \cellcolor{BaselineRow}\underline{0.4824} & \cellcolor{BaselineRow}0.5035 & \cellcolor{BaselineRow}0.5020 & \cellcolor{BaselineRow}0.6590 & \cellcolor{BaselineRow}0.6521 & \cellcolor{BaselineRow}0.5489 & \cellcolor{BaselineRow}0.5455 \\
 & \cellcolor{BaselineRow}Attribution & \cellcolor{BaselineRow}0.4779 & \cellcolor{BaselineRow}0.4810 & \cellcolor{BaselineRow}0.5020 & \cellcolor{BaselineRow}\underline{0.5082} & \cellcolor{BaselineRow}\underline{0.6648} & \cellcolor{BaselineRow}\underline{0.6637} & \cellcolor{BaselineRow}0.5482 & \cellcolor{BaselineRow}\underline{0.5510} \\
 & \cellcolor{BaselineRow}Diversity & \cellcolor{BaselineRow}0.4817 & \cellcolor{BaselineRow}0.4690 & \cellcolor{BaselineRow}\textbf{0.5224} & \cellcolor{BaselineRow}0.4965 & \cellcolor{BaselineRow}0.6543 & \cellcolor{BaselineRow}0.6477 & \cellcolor{BaselineRow}0.5528 & \cellcolor{BaselineRow}0.5377 \\
 & \cellcolor{BaselineRow}Attr-Div & \cellcolor{BaselineRow}0.4855 & \cellcolor{BaselineRow}0.4820 & \cellcolor{BaselineRow}0.5027 & \cellcolor{BaselineRow}0.4933 & \cellcolor{BaselineRow}0.6568 & \cellcolor{BaselineRow}0.6492 & \cellcolor{BaselineRow}0.5483 & \cellcolor{BaselineRow}0.5415 \\
  & \cellcolor{BaselineRow}TSDS & \cellcolor{BaselineRow}\textbf{0.4939} & \cellcolor{BaselineRow}0.4803 & \cellcolor{BaselineRow}0.5145 & \cellcolor{BaselineRow}0.5067 & \cellcolor{BaselineRow}\textbf{0.6656} & \cellcolor{BaselineRow}0.6575 & \cellcolor{BaselineRow}\textbf{0.5580} & \cellcolor{BaselineRow}0.5481 \\
 & \cellcolor{OursRow}\textbf{\method} & \cellcolor{OursRow}\underline{0.4894} & \cellcolor{OursRow}\textbf{0.4863} & \cellcolor{OursRow}\underline{0.5153} & \cellcolor{OursRow}\textbf{0.5161} & \cellcolor{OursRow}0.6645 & \cellcolor{OursRow}\textbf{0.6641} & \cellcolor{OursRow}\underline{0.5564} & \cellcolor{OursRow}\textbf{0.5555} \\
\midrule
\bottomrule
\end{tabular}
}%
\caption{Performance comparison on biomedical/healthcare datasets (\texttt{3b} base model). \textbf{Bold} and \underline{underscore} denote top-1/2 accuracy.}
\label{tab:strong-table-3}
\end{table*}

\subsection{Main Results}
\label{ssec:exp-results}
In this subsection, we present the main experiment results by comparing \method\ with other baseline data selection methods across various settings. 

\paragraph{\method\ outperforms baseline selection methods.}

We present our main results in Table~\ref{tab:strong-table-1}, \ref{tab:strong-table-2} and \ref{tab:strong-table-3}, which are conducted under the \textit{strong generator setting} (i.e. \texttt{3B} base model and \texttt{14B} data generator). Across all datasets, \method\ mostly outperforms other baseline methods under different selection ratios and iterations.  The experiment results conducted under the \textit{weak generator setting} (i.e. \texttt{14B} base model and \texttt{14B} data generator) are deferred to Appendix~\ref{app:weak-exp} due to space limit. We observe that the trend follows even when the data generator is not strictly stronger than the base model. These results indicate that \method\ identifies data that is \textit{fundamentally} useful, i.e., data that remains beneficial for LLM adaptation regardless of the generator capacity or selection ratio.

\paragraph{\method\ consistently improves base LLM.}

\begin{figure}[t]
\centering
\includegraphics[width=1.0\linewidth]{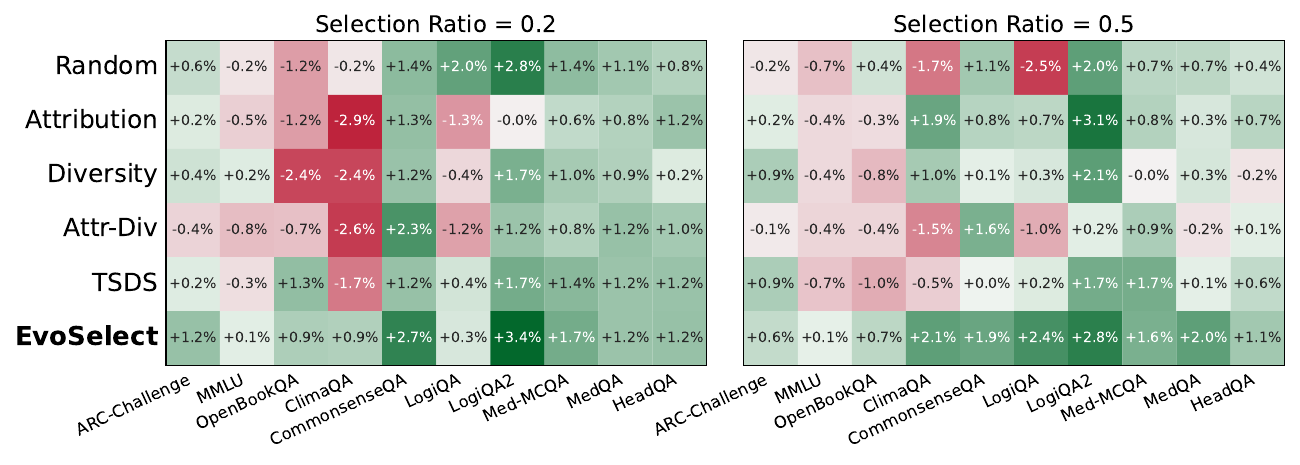}
\caption{
Performance gain over \texttt{3b} base model. The figure shows that only \method\ has consistent improvement across selection ratios \& datasets.}
\label{fig:heatmap-over-base}
\end{figure}

In Figure~\ref{fig:heatmap-over-base}, we measure the performance difference between each method's average performance (over two iterations) and that of the base model. We show that \method\ is the \textit{only} selection method that constantly outperforms base model performance (\textcolor{teal}{\textit{green}} indicates beneficial adaption and \textcolor{red}{\textit{red}} indicates harmful adaption). For all other baselines, there are certain failure cases, indicating their unstable LLM adaption efficacy which can even harm the model performance.

\subsection{Discussion}
\label{ssec:discussion}

In this subsection, we provide further discussion to analyze \method's performance and its difference against prior work. More discussion is deferred to Appendix~\ref{app:further-analysis}. 

\paragraph{\method's performance gain vs base LLM capacity.}

\begin{figure}[ht]
\centering
\includegraphics[width=1.0\linewidth]{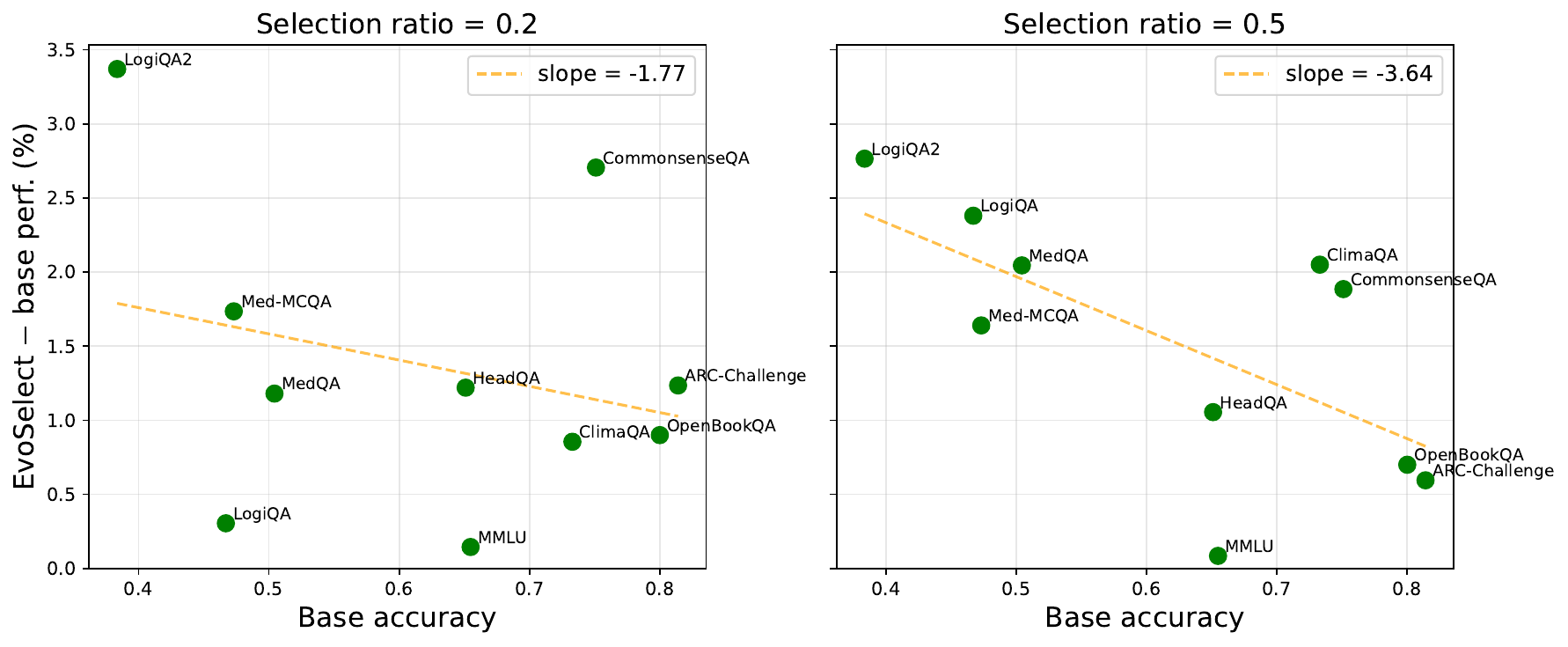}
\caption{Performance gain vs task difficulty. \method's advantage becomes more significant as task difficulty increases (i.e. when base perf. is lower).}
\label{fig:gain-vs-base-acc}
\end{figure}

One interesting observation is that \method's relative performance against the base model increases as task difficulty grows. As shown in Figure~\ref{fig:gain-vs-base-acc}, the performance gain becomes more pronounced on more challenging tasks, suggesting that \method\ is particularly effective in scenarios where stronger adaptation is required.

\paragraph{\method\ vs TSDS~\citep{liu2024tsds}.} 

We describe the key differences between TSDS~\citep{liu2024tsds} and \method\ in this paragraph. 
% TSDS optimizes a transport plan from validation to training examples and derives a sampling distribution, whereas \method\ directly optimizes a weight distribution over training samples via iterative updates, enabling a global viewpoint. 
While TSDS concatenates target alignment and diversity enhancement, we note that it has several limitations: (1) its diversity is enforced through row-wise transport, which operates locally for each validation example and does not explicitly prevent redundancy among selected samples; (2) its solution effectively reduces to a nearest-neighbor-style selection, limiting its expressiveness to local matching without capturing global interactions among training samples. The decoupled treatment of the two objectives ultimately leads to suboptimal training sets, resulting in inferior performance compared to \method, as demonstrated in Section~\ref{ssec:exp-results}.

%% file: sections/05Related.tex
\section{Related Work}

\paragraph{Data Selection for LLM Adaptation.} Existing work can be grouped into attribution, diversity and hybrid approaches. Attribution-only methods~\citep{xia2024less,san2024in2core,zhou2024hyperinf,kwon2023datainf,cao2023instruction} mostly rely on variants of influence-function~\citep{koh2017understanding} to estimate the contribution of each sample, but they may under-cover the target distribution. Diversity-only methods~\citep{wu2023self,wang2024diversity,jung2025prismatic} instead improve coverage via sampling/clustering, but they do not model task alignment. Some works attempt to consider both aspects: in pretraining, these methods~\citep{liu2025quadmix,zhuang2025meta,wang2025tikmix} combine quality and diversity through mixture weighting. However, these methods cannot be directly applied to target task adaption since (1) they operate without target-conditioned signals and (2) they learn mixture weights at the corpus or domain level instead of sample level. For targeted fine-tuning, some methods~\citep{zhang2025d3, wu2024best} typically combine multiple criteria through heuristic scoring functions and TSDS~\citep{liu2024tsds} solves a distribution alignment problem with an additional diversity regularizer and performs sampling to select data. In summary, these methods either rely solely on attribution/diversification or combine the two in a heuristic manner.

\paragraph{Optimal Transport for Data Attribution.}
We note that a line of work leverages Optimal Transport (OT)~\citep{peyre2019computational} to compute data attribution scores.  LAVA~\citep{just2023lava} and SAVA~\citep{kessler2024sava} directly use the magnitude of the calibrated gradient of the OT objective as attribution scores. 
In contrast, TSDS~\citep{liu2024tsds} solves for a probability distribution over candidate data via an OT-based objective with local diversity augmentation, and samples training instances from this distribution. Similarly, TAROT~\citep{feng2024tarot} leverages greedy algorithm to select data that minimizes OT distance between candidate training and validation samples. We note that to the best of our knowledge, \method\ is the first OT-based method that directly synergizes attribution and diversity objective for targeted data selection.

% \paragraph{Diversity-based Data selection for LLM Adaptation.}

% overall setting:

% ancient: sTar
% recently: spend wisely

% Improving Influence-based Instruction Tuning Data Selection for Balanced
% Learning of Diverse Capabilities

% LEAD: Iterative Data Selection for Efficient LLM Instruction Tuning

%% file: sections/06Conclusion.tex
\section{Conclusion}

% We studied the problem of data-efficient adaptation of LLM under synthetic data generation, where the primary challenge is that not all generated examples contribute equally to learning. We proposed \method, an iterative framework that combines data attribution and diversification within an OT-based selection procedure to identify high-utility training examples for target tasks. Across reasoning and knowledge-intensive benchmarks, \method\ consistently outperforms other selection baselines, including training on all generated data and heuristic selection strategies, demonstrating that principled data selection is critical for effective and efficient LLM adaptation. \tw{need update}

We study data-efficient LLM adaptation under an \textit{iterative generation-selection-training loop}, where generated samples can be redundant or misaligned with the target task and thus need to be selected before training. We propose \method, an OT-based selection framework that jointly models task alignment and diversity to identify high-utility training examples for each iteration. Experiments across various benchmarks show that \method\ consistently outperforms existing baselines under both weak and strong generators, highlighting the importance of a principled selection framework to achieve effective LLM adaptation.

\section*{Acknowledgement}
\thanks{This work is supported by NSF (2416070). The content of the information in this document does not necessarily reflect the position or the policy of the Government, and no official endorsement should be inferred.  The U.S. Government is authorized to reproduce and distribute reprints for Government purposes notwithstanding any copyright notation here on.
}

%% file: sections/99Appendix.tex
\section{Additional Preliminary}
\label{app:addition-prelim}

\paragraph{Optimal Transport.}
Optimal transport (OT)~\citep{peyre2019computational,flamary2021pot} provides a principled way to measure the discrepancy between two probability distributions. Let $(\mathcal{Z}, d)$ be a metric space equipped with a ground cost function $d: \mathcal{Z}\times\mathcal{Z}\to\mathbb{R}_{\geq 0}$, where $d(z,z')$ denotes the distance between points $z, z'\in\mathcal{Z}$. Given two probability measures $\mu$ and $\nu$ defined on $\mathcal{Z}$, the optimal transport distance between $\mu$ and $\nu$ is defined as: $\operatorname{OT}(\mu,\nu,d) \;=\; \inf_{\pi\in\Pi(\mu,\nu)} 
\int_{\mathcal{Z}\times\mathcal{Z}} d(z,z')\,\mathrm{d}\pi(z,z')$,
where $\Pi(\mu,\nu)$ is the set of all couplings over $\mathcal{Z}\times\mathcal{Z}$ with marginals $\mu$ and $\nu$. When $\mu$ and $\nu$ are \emph{empirical distributions} supported on finite sample sets,
$\mu=\frac{1}{n}\sum_{i=1}^n\delta_{z_i},
\nu=\frac{1}{m}\sum_{j=1}^m\delta_{z'_j},$
where $\delta$ denotes the Dirac measure, the OT problem reduces to a discrete linear program. Let $M\in\mathbb{R}^{n\times m}$ be the cost matrix with entries $M_{ij}=d(z_i,z'_j)$; then the OT distance can be expressed as: $\operatorname{OT}(\mu,\nu,M) \;=\; 
\min_{\pi\in\Pi(\mu,\nu)} \sum_{i=1}^n\sum_{j=1}^m M_{ij}\,\pi_{ij}$,
where $\pi_{ij}$ denotes the amount of mass transported from $z_i$ to $z'_j$ under the coupling $\pi$.

\section{Dataset Settings}
\label{app:dataset-settings}
% \tw{one table: train/val/test partition for each dataset and dataset stats }

We detail the data split for seed, validation and test samples as following. We randomly obtain 100 samples from the seed split as initial seed and 1000 samples from the validation split as validation data for targeted task alignment. The full test split is used for evaluation. For each dataset, all three splits are disjoint.

\begin{table}[ht]
\centering
\begin{tabular}{c|cccc}
\toprule
\midrule
 Dataset Name & \texttt{ARC-Challenge} & \texttt{MMLU} & \texttt{OpenBookQA} & \texttt{ClimaQA} \\
 \midrule
 Seed Split & Train & Dev & Train &  Silver \\
 Validation Split & Val & Val &  Val & Silver  \\
 Test Split  & Test & Test & Test & Gold\\
 \midrule
 \bottomrule
\end{tabular}
\caption{Data split setting for scientific datasets.}
\label{app:tab-split-1}
\end{table}

\begin{table}[ht]
\centering
\begin{tabular}{c|ccc}
\toprule
\midrule
 Dataset Name & \texttt{CommonsenseQA} & \texttt{LogiQA}  & \texttt{LogiQA2}  \\
 \midrule
 Seed Split & Train &Train  &  Train    \\
 Validation Split& Train & Train  & Train   \\
 Test Split & Val &Val  & Test \\
 \midrule
 \bottomrule
\end{tabular}
\caption{Data split setting for commonsense/logical  datasets.}
\label{app:tab-split-2}
\end{table}

\begin{table}[ht]
\centering
\begin{tabular}{c|ccc}
\toprule
\midrule
 Dataset Name & \texttt{Med-MCQA} & \texttt{MedQA}& \texttt{HeadQA} \\
 \midrule
 Seed Split & Train & Train &  Train    \\
 Validation Split & Train & Train &  Val  \\
 Test Split & Val & Test & Test \\
 \midrule
 \bottomrule
\end{tabular}
\caption{Data split setting for biomedical/healthcare datasets.}
\label{app:tab-split-3}
\end{table}

\section{Synthetic Training Data Generation Details and Prompts}
\label{app:gen-prompts}
We mainly adopt the settings given by \citet{jung2025prismatic}. In the following, we show the detailed setting \& prompts for question and solution generation, correspondingly.

% prompts we use for (1) question generation with few-shot examples sampled from seed set and (2) solution generation (system prompt and instruction prompt).

\paragraph{Question generation.}
For question generation, we adopt the few-shot setting, where 5 random few-shot exampled are extracted from the current seed set and augmented to the prompt for novel question synthesis. At each iteration, 5k problem instance is generated. For text generation setting, the temperature is set as $1.0$, $\text{top\_p}=0.95$ and $\text{max\_tokens} = 8192$. The prompt is as follows.

\noindent
\begin{promptframe}
\begin{lstlisting}[
  basicstyle=\ttfamily\small,
  breaklines=true,
  breakatwhitespace=true,
  keepspaces=true,
  xleftmargin=0pt,
  breakindent=0pt
]
Given a set of example multiple-choice questions, create 1 new multiple-choice questions that are strictly harder and test deeper understanding.

Each question must have exactly 4 options labeled (A), (B), (C), (D), with exactly one correct answer.

Format each question as:

---
[[Problem]]
<Question text>

(A) <option A>
(B) <option B>
(C) <option C>
(D) <option D>

[[Answer]]
<correct letter, e.g., A>
---

Examples:
[[Problem]]
$#$problem$#$

[[Answer]]
$#$answer$#$
\end{lstlisting}
\end{promptframe}

\paragraph{Solution generation.}

For solution generation, we adopt the widely known \textit{rationale generation}~\citep{zelikman2022star} to aid solution generation. Specifically, the data generator is prompted to provide a 1-2 sentence rationale before answering. For text generation setting, the temperature is set as $0.75$, $\text{top\_p}=0.95$ and $\text{max\_tokens} = 2048$. Each question, we will generate 3 solutions. To obtain seed and validation set, we use the ground-truth solution to filter incorrect ones; while when generating  candidate training samples, we use majority-voting to preserve  confident rationales \& answers. The system and instruction prompts are as follows.

% adopt the few-shot setting, where 5 random few-shot exampled are extracted from the current seed set and augmented to the prompt for novel question synthesis. 

\begin{promptframe}
\begin{lstlisting}[
  basicstyle=\ttfamily\small,
  breaklines=true,
  breakatwhitespace=true,
  keepspaces=true,
  xleftmargin=0pt,
  breakindent=0pt
]
<System Prompt>
You are an expert at answering multiple-choice questions. Your role is to:
1. Carefully analyze the question and all options
2. Provide a concise 1-2 sentence rationale explaining your reasoning
3. Select the correct answer with confidence

Always provide a 1-2 sentence brief explanation before giving the final answer.

<Instruction Prompt>
Answer the following multiple-choice question by selecting the correct option.

Provide a concise 1-2 sentence rationale explaining your reasoning, then give your answer in the format: Answer: [LETTER], where [LETTER] is one of A, B, C, D.

$#$problem$#$

\end{lstlisting}
\end{promptframe}

% \tw{prompt and generation configuration for problem/answer generation.}

% For each question, we query the data generator to 
% \tw{how to get train and val, ground-truth filtering, how many per question}

\section{Training Configurations}
\label{app:training-config}
The base model is post-trained using SFT with LoRA~\citep{hu2022lora} adapters through TRL~\citep{vonwerra2020trl}. For LoRA, we set $r=64$, $\alpha=128$ and dropout rate $0.05$. LoRA is applied to all linear layers. The per-device batch size is $4$ with gradient accumulation of $4$. For each iteration, the base models are trained for $3$ epochs with a constant learning rate of $2\times10^{-5}$, weight decay $10^{-4}$, and warm-up ratio $0.05$. The best checkpoint is selected based on validation loss. Our experiments are conducted on an Ubuntu server equipped with Intel Xeon Platinum x86-64 CPUs and 4 Nvidia A100 80GB GPUs.

\section{Evaluation Details and Prompts}
\label{app:evaluation}
We adopt a popular evaluation pipeline, Evalscope~\citep{evalscope_2024}, for all datasets. For \texttt{ARC-Challenge}, \texttt{MMLU}, \texttt{CommonsenseQA}, \texttt{LogiQA} and \texttt{Med-MCQA}, we use the built-in protocol; for the remaining ones, we download the datasets from Huggingface~\citep{huggingface}. For text generation setting, the temperature is set as $0.0$ and $\text{max\_tokens}= 1024$. For consistency, we adopt zero-shot evaluation for all datasets. The system and instruction prompts are as follows.

\begin{promptframe}
\begin{lstlisting}[
  basicstyle=\ttfamily\small,
  breaklines=true,
  breakatwhitespace=true,
  keepspaces=true,
  xleftmargin=0pt,
  breakindent=0pt
]
<System Prompt>
You are Qwen, created by Alibaba Cloud. You are a helpful assistant.

<Instruction Prompt>
Answer the following multiple choice question by selecting the correct option.

Provide a concise 1-2 sentence rationale explaining your reasoning, then give your answer in the format: Answer: [LETTER] where [LETTER] is one of {letters}.
Question: {question}
Options: {choices}

\end{lstlisting}
\end{promptframe}

% evaluation details (evalscope) 
% zero-shot, temp

% and prompts

\section{Baselines}
\label{app:baselines}
% \vspace{-2mm}
We elaborate the detailed setup for each baseline we compare in Section~\ref{sec:exp}.
\begin{enumerate}
    \item \textbf{Attribution}~\citep{xia2024less}  selects top-$k$ samples with highest scores, which are computed by the inner product between training and the average validation representations. Gradients are obtained from proxy model and no warm-up training is involved. 
    \item \textbf{Diversity}~\citep{jung2025prismatic} first runs K-means clustering over the training representations, then keeps filling the candidate pool from the smallest clusters until the budget is exhausted. We follow the default configuration: cluster\_ratio = 0.1 (i.e. \#clusters = \#samples * cluster\_ratio).
    \item \textbf{Attr-Div} first filters out all the bottom-$25\%$ data by attribution scores, then apply the same selection strategy as \textbf{Diversity} to select data from the pruned candidates.
    \item \textbf{TSDS}~\citep{liu2024tsds} finds candidate samples that are closest to the validation examples via OT and down-weights redundant ones using a kernel density estimate (KDE). It then samples from the associated probability distribution. We follow the default configuration: max\_k (maximum kNN neighbors) is set to 5000, kde\_k (neighborhood size for KDE) is set to 1000, KDE bandwidth $\sigma=0.75$, OT regularization $\alpha=0.5$, OT regularization scale  $C=5.0$.
\end{enumerate}

    % Kernel density estimation (KDE) over the candidate neighborhood downweights near-duplicate training samples before selection, promoting diversity within the transport solution.

% TSDS: Transport-based Sample Data Selection frames data selection as an optimal transport problem. For each validation query, the k nearest training candidates (in gradient space) are retrieved. A heap-based algorithm iteratively assigns probability mass across candidates, governed by a budget parameter  that trades off coverage against concentration. Kernel density estimation (KDE) over the candidate neighborhood downweights near-duplicate training samples before selection, promoting diversity within the transport solution.

\section{Additional Experiment Results}
\label{app:weak-exp}
\paragraph{Main results with weak data generator.}

In Table~\ref{tab:weak-table-1}, \ref{tab:weak-table-2} and \ref{tab:weak-table-3}, we show that with weak data generator (i.e. \texttt{14B} data generator), \method\ is the most effective across settings.

% \newpage
\input{tables/weak_big_tables}

\paragraph{Different backbone model family.}

We adapt our experiments to the Llama-3 family~\citep{grattafiori2024llama}. The proxy model is \texttt{Llama-3.2-1B-Instruct}, the data generator is \texttt{Llama-3.1-8B-Instruct} and the base model is \texttt{Llama-3.2-3B-Instruct}. We evaluate methods on one representative dataset from each domain with selection ratio 0.2. We consider random and TSDS as main baselines (strongest across all studied benchmarks). In Table~\ref{tab:different-backbone}, we show that \method\ is still effective when adapted to a different model family.

\begin{table}[h]
\centering
\begin{tabular}{c|c|ccc}
\toprule
\midrule
Iteration & Method & \texttt{ARC-Challenge} & \texttt{CommonsenseQA} & \texttt{HeadQA} \\
\midrule
Iter 0 & Base      & 0.7389 & 0.7281 & 0.6247 \\
\midrule
\multirow{3}{*}{Iter 1}
       & \method & \textbf{0.7611} & \textbf{0.7396} & \textbf{0.6568} \\
       & Random    & 0.7201          & 0.7322          & 0.6459 \\
       & TSDS      & 0.7398          & 0.7361          & 0.6368 \\
\midrule
\multirow{3}{*}{Iter 2}
       & \method & \textbf{0.7799} & 0.7256          & \textbf{0.6499} \\
       & Random    & 0.7594          & \textbf{0.7338} & 0.6386 \\
       & TSDS      & 0.7457          & 0.7273          & 0.6368 \\
\midrule
\bottomrule
\end{tabular}
\caption{Performance across three representative downstream benchmarks with model backbones/proxies from the Llama-3 family.}
\label{tab:different-backbone}
\end{table}

\paragraph{Different proxy model family.}

To validate the cross-family transferability, we use \texttt{Llama-3.2-1B-Instruct} as the proxy model for gradient computation and the base model/data generator are \texttt{Qwen2.5-3B-Instruct}/\texttt{Qwen2.5-14B-Instruct}, respectively. We evaluate methods on one representative dataset from each domain with selection ratio 0.2. We consider random and TSDS as main baselines (strongest across all studied benchmarks). In Table~\ref{tab:different-proxy}, we can observe that \method\ outperforms across most settings.

\begin{table}[ht]
\centering
\begin{tabular}{c|c|ccc}
\toprule
\midrule
Iteration & Method & \texttt{ARC-Challenge} & \texttt{CommonsenseQA} & \texttt{HeadQA} \\
\midrule
\multirow{3}{*}{Iter 1}
       & \method & \textbf{0.8311} & \textbf{0.7854} & 0.6597 \\
       & Random    & 0.8140          & 0.7674          & 0.6594 \\
       & TSDS      & 0.8097          & 0.7559          & \textbf{0.6605} \\
\midrule
\multirow{3}{*}{Iter 2}
       & \method & \textbf{0.8242} & 0.7731          & \textbf{0.6619} \\
       & Random    & 0.8123          & \textbf{0.7756} & 0.6543 \\
       & TSDS      & 0.8131          & 0.7740          & 0.6557 \\
\midrule
\bottomrule
\end{tabular}
\caption{Performance across three representative downstream benchmarks with cross-model backbones \& proxies.}
\label{tab:different-proxy}
\end{table}

\paragraph{More iterations.}

We extend the experiments to 5 iterations (Iteration 1 \& 2 omitted). We evaluate methods on one representative dataset from each domain with selection ratio 0.2. We consider random and TSDS as main baselines (strongest across all studied benchmarks). In Table~\ref{tab:more-iterations}, we observe that (i) \method\ achieves the best performance across most settings and (ii) \method\ can progressive improve over more iterations.

\begin{table}[ht]
\centering
\begin{tabular}{c|c|ccc}
\toprule
\midrule
Iteration & Method & \texttt{ARC-Challenge} & \texttt{CommonsenseQA} & \texttt{HeadQA} \\
\midrule
Iter 0 & Base & 0.8140 & 0.7510 & 0.6510 \\
\midrule
\midrule
\multirow{3}{*}{Iter 3}
       & \method & \textbf{0.8225} & \textbf{0.7821} & 0.6550 \\
       & Random  & 0.8097          & 0.7715          & \textbf{0.6590} \\
       & TSDS    & 0.8157          & 0.7641          & 0.6572 \\
\midrule
\multirow{3}{*}{Iter 4}
       & \method & \textbf{0.8268} & \textbf{0.7879} & 0.6543 \\
       & Random  & 0.8245          & 0.7772          & 0.6411 \\
       & TSDS    & 0.8140          & 0.7600          & \textbf{0.6561} \\
\midrule
\multirow{3}{*}{Iter 5}
       & \method & \textbf{0.8370} & \textbf{0.7903} & \textbf{0.6616} \\
       & Random  & 0.8217          & 0.7797          & 0.6586 \\
       & TSDS    & 0.8063          & 0.7641          & 0.6539 \\
\midrule
\bottomrule
\end{tabular}
\caption{Performance across three representative downstream benchmarks with 5 iterations.}
\label{tab:more-iterations}
\end{table}

\section{Further Analysis}
\label{app:further-analysis}

\paragraph{Ablation study of \method.}

We evaluate \method\ and its variant on one representative dataset from each domain with selection ratio 0.2. The variants are as follows.
\begin{itemize}
    \item \textbf{\method-OT-only}: \method\ without the diversity-enhancing term. This can also be considered as a proxy of the TAROT~\citep{feng2024tarot} method.
    \item \textbf{\method-no-std}: \method\ without the standarization step.
    \item \textbf{\method-40-step}: \method\ with more optimization steps.
\end{itemize}

In Table~\ref{tab:method-ablation}, we can observe that \method\ outperforms other variants across most settings. Also, \method's performance is stable across different numbers of steps.

\begin{table}[ht]
\centering
\begin{tabular}{c|c|ccc}
\toprule
\midrule
Iteration & Method & \texttt{ARC-Challenge} & \texttt{CommonsenseQA} & \texttt{HeadQA} \\
\midrule
\multirow{7}{*}{Iter 1}
       & \method-10-step & 0.8208          & 0.7772          & 0.6630 \\
       & \method-20-step & \textbf{0.8285} & 0.7748          & 0.6565 \\
       & \method-40-step & 0.8131          & 0.7821          & \textbf{0.6652} \\
       & \method-no-std  & 0.8157          & \textbf{0.7887} & 0.6616 \\
       & \method-OT-only & 0.8168          & 0.7621          & 0.6623 \\
       & Div-only        & 0.8174          & 0.7723          & 0.6524 \\
       & TSDS            & 0.8063          & 0.7707          & 0.6601 \\
\midrule
\multirow{7}{*}{Iter 2}
       & \method-10-step & 0.8106          & 0.7797          & 0.6503 \\
       & \method-20-step & \textbf{0.8285} & 0.7862          & 0.6645 \\
       & \method-40-step & 0.8106          & \textbf{0.7903} & 0.6528 \\
       & \method-no-std  & 0.8191          & 0.7723          & 0.6539 \\
       & \method-OT-only & 0.8131          & 0.7879          & 0.6501 \\
       & Div-only        & 0.8191          & 0.7609          & 0.6543 \\
       & TSDS            & 0.8229          & 0.7559          & \textbf{0.6656} \\
\midrule
\bottomrule
\end{tabular}
\caption{Ablation studies conducted across three representative downstream benchmarks.}
\label{tab:method-ablation}
\end{table}

\paragraph{Empirical runtime analysis.}

We record empirical number of running time in Table~\ref{tab:runtime}. \method\ adds negligible computation overhead to the overall pipeline.

\begin{table}[ht]
\centering
\begin{tabular}{l|r}
\toprule
\midrule
Procedure & Time (s) \\
\midrule
Data generation            & 1628.81 \\
Model training             & 1252.44 \\
Proxy gradient collection  & 307.23  \\
\midrule
\multicolumn{2}{c}{Selection methods (per iteration)} \\
\midrule
Random      & 0.004 \\
Attribution & 0.031 \\
Diversity   & 0.085 \\
TSDS        & 6.98  \\
\method     & 7.46  \\
\midrule
\bottomrule
\end{tabular}
\caption{Runtime of each component within the pipeline. \method\ is comparable against baselines with OT computation. }
\label{tab:runtime}
\end{table}

\begin{figure}[ht]
\centering
\includegraphics[width=1.0\linewidth]{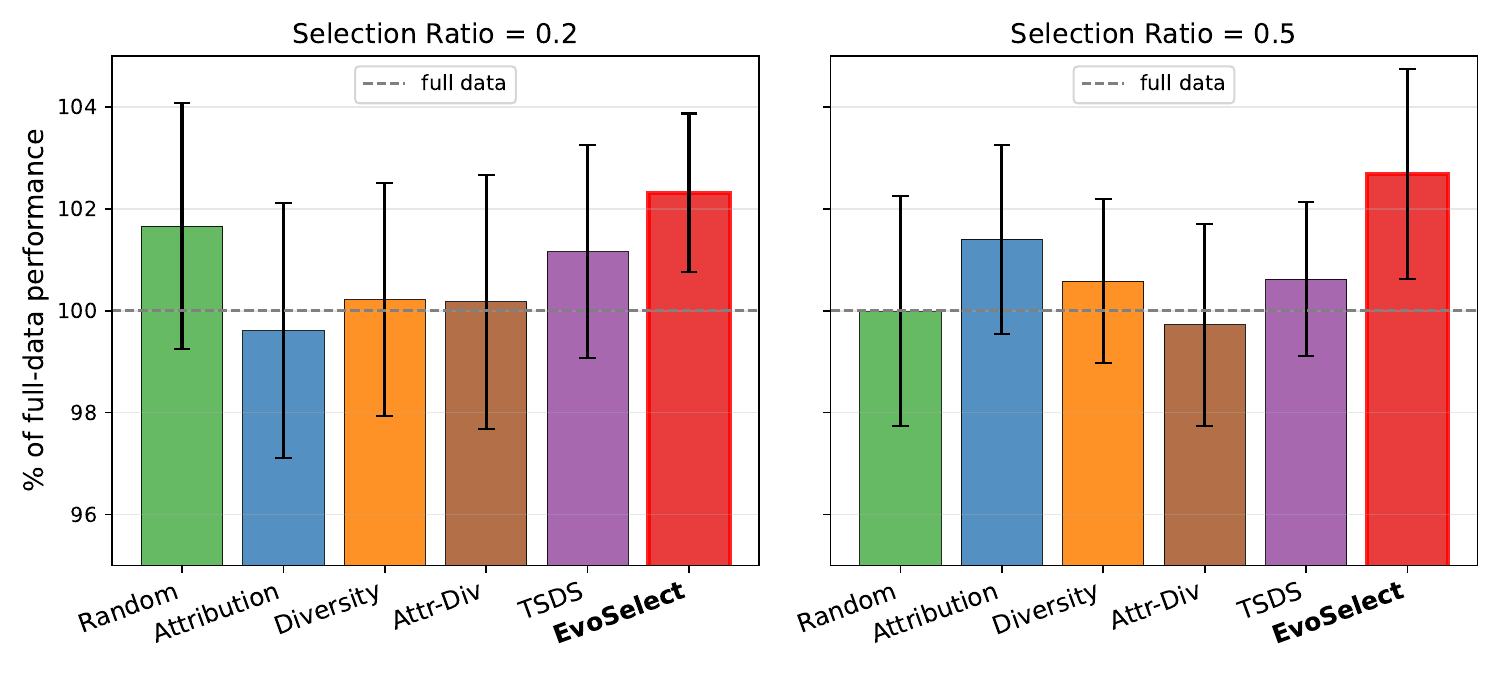}
\caption{
Performance relative to full-data training. \method\ consistently achieves the highest performance across both selection ratios, exceeding the full-data baseline.
}
\label{fig:gain-vs-full-data}
\end{figure}

\paragraph{Performances vs full-data training.}
Figure~\ref{fig:gain-vs-full-data} compares \method\ against baseline methods relative to training on the full dataset under two selection ratios by enumerating performances across all datasets.  Across both settings, \method\ consistently achieves the highest performance, surpassing all baselines.  Notably, even at a lower selection ratio, \method\ consistently exceeds full-data performance, indicating its ability to identify high-quality subsets. In contrast, attribution-only and diversity-only methods exhibit inconsistent behavior, often falling below the full-data baseline, while heuristic combinations (\textit{Attr-Div}) provide limited improvement. TSDS~\citep{liu2024tsds} shows competitive performance but remains consistently below \method, suggesting that its way of combining alignment and diversity is not sufficiently effective.

\begin{figure}[ht]
\centering
\includegraphics[width=1.0\linewidth]{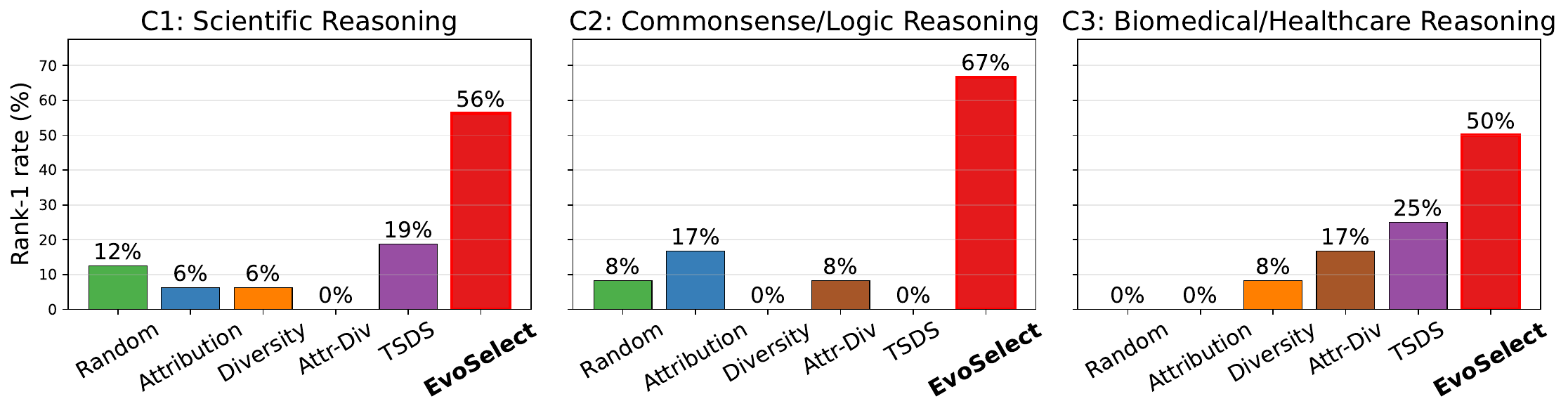}
\caption{
Win-rate across task domains. \method\ attains the highest rank-1 rate across all task domains, demonstrating robust top-performing behavior across different task domains.
}
\label{fig:winrate}
\end{figure}

\paragraph{Win-rate across task domains.}
Figure~\ref{fig:winrate} reports the rank-1 win rate across different task domains. \method\ achieves the highest win rate in all domains. This indicates that \method\ consistently identifies superior training subsets across diverse domains. In contrast, attribution and diversity-based methods exhibit low and unstable win rates, while TSDS achieves moderate performance but lacks consistency across domains. These results highlight the robustness of \method\ in selecting effective data across varying task characteristics.

\section{Additional Related Work}

While this paper mainly study data-side diversity, we acknowledge that there is also a line of work studying generator-side diversity~\citep{yu2023large,divekar2024synthesizrr,chen2024diversity, kowshik2024corrsynth,ge2024scaling}.

%% file: tables/weak_big_tables.tex
\begin{table*}[!htbp]
\centering
\renewcommand{\arraystretch}{0.75}
\resizebox{1.0\textwidth}{!}{%
\begin{tabular}{c l cc cc cc cc cc}
\toprule
\midrule
 &  & \multicolumn{2}{c}{\textbf{ARC-Challenge}} & \multicolumn{2}{c}{\textbf{MMLU}} & \multicolumn{2}{c}{\textbf{OpenBookQA}} & \multicolumn{2}{c}{\textbf{ClimaQA}} & \multicolumn{2}{c}{\textbf{\textbf{Avg.}}} \\
\cmidrule(lr){3-4} \cmidrule(lr){5-6} \cmidrule(lr){7-8} \cmidrule(lr){9-10} \cmidrule(lr){11-12}
 & \textbf{Selection Ratio} $\rightarrow$ & \multicolumn{1}{c}{0.2} & \multicolumn{1}{c}{0.5} & \multicolumn{1}{c}{0.2} & \multicolumn{1}{c}{0.5} & \multicolumn{1}{c}{0.2} & \multicolumn{1}{c}{0.5} & \multicolumn{1}{c}{0.2} & \multicolumn{1}{c}{0.5} & \multicolumn{1}{c}{0.2} & \multicolumn{1}{c}{0.5} \\
\cmidrule{2-12}
 & \textbf{Methods} $\downarrow$ & & & & & & & & & & \\
\cmidrule{2-12}
 & \cellcolor{BaseRow}Base & \multicolumn{2}{c}{\cellcolor{BaseRow}0.9181} & \multicolumn{2}{c}{\cellcolor{BaseRow}0.7882} & \multicolumn{2}{c}{\cellcolor{BaseRow}0.9140} & \multicolumn{2}{c}{\cellcolor{BaseRow}0.7945} & \multicolumn{2}{c}{\cellcolor{BaseRow}0.8537} \\
\midrule
\multirow{7}{*}{\textbf{Iter.\ 1}} & \cellcolor{AllRow}All & \multicolumn{2}{c}{\cellcolor{AllRow}0.9164} & \multicolumn{2}{c}{\cellcolor{AllRow}0.7911} & \multicolumn{2}{c}{\cellcolor{AllRow}0.9260} & \multicolumn{2}{c}{\cellcolor{AllRow}0.7979} & \multicolumn{2}{c}{\cellcolor{AllRow}0.8579} \\
 & \cellcolor{BaselineRow}Random & \cellcolor{BaselineRow}\underline{0.9189} & \cellcolor{BaselineRow}\textbf{0.9181} & \cellcolor{BaselineRow}\underline{0.7922} & \cellcolor{BaselineRow}0.7879 & \cellcolor{BaselineRow}\underline{0.9300} & \cellcolor{BaselineRow}\underline{0.9340} & \cellcolor{BaselineRow}0.8048 & \cellcolor{BaselineRow}0.8014 & \cellcolor{BaselineRow}0.8615 & \cellcolor{BaselineRow}0.8604 \\
 & \cellcolor{BaselineRow}Attribution & \cellcolor{BaselineRow}0.9138 & \cellcolor{BaselineRow}\textbf{0.9181} & \cellcolor{BaselineRow}0.7919 & \cellcolor{BaselineRow}\underline{0.7906} & \cellcolor{BaselineRow}\textbf{0.9440} & \cellcolor{BaselineRow}0.9320 & \cellcolor{BaselineRow}\textbf{0.8219} & \cellcolor{BaselineRow}0.7911 & \cellcolor{BaselineRow}\textbf{0.8679} & \cellcolor{BaselineRow}0.8579 \\
 & \cellcolor{BaselineRow}Diversity & \cellcolor{BaselineRow}0.9181 & \cellcolor{BaselineRow}0.9147 & \cellcolor{BaselineRow}0.7886 & \cellcolor{BaselineRow}0.7875 & \cellcolor{BaselineRow}0.9220 & \cellcolor{BaselineRow}0.9320 & \cellcolor{BaselineRow}0.8048 & \cellcolor{BaselineRow}\textbf{0.8116} & \cellcolor{BaselineRow}0.8584 & \cellcolor{BaselineRow}\underline{0.8614} \\
 & \cellcolor{BaselineRow}Attr-Div & \cellcolor{BaselineRow}\textbf{0.9249} & \cellcolor{BaselineRow}\underline{0.9172} & \cellcolor{BaselineRow}0.7891 & \cellcolor{BaselineRow}0.7886 & \cellcolor{BaselineRow}0.9260 & \cellcolor{BaselineRow}0.9220 & \cellcolor{BaselineRow}0.8116 & \cellcolor{BaselineRow}\textbf{0.8116} & \cellcolor{BaselineRow}0.8629 & \cellcolor{BaselineRow}0.8599 \\
  & \cellcolor{BaselineRow}TSDS & \cellcolor{BaselineRow}0.9300 & \cellcolor{BaselineRow}0.9280 & \cellcolor{BaselineRow}0.7903 & \cellcolor{BaselineRow}0.7875 & \cellcolor{BaselineRow}0.9220 & \cellcolor{BaselineRow}0.9240 & \cellcolor{BaselineRow}\underline{0.8185} & \cellcolor{BaselineRow}\underline{0.8185} & \cellcolor{BaselineRow}0.8652 & \cellcolor{BaselineRow}0.8645 \\
 & \cellcolor{OursRow}\textbf{\method} & \cellcolor{OursRow}\underline{0.9189} & \cellcolor{OursRow}\textbf{0.9181} & \cellcolor{OursRow}\textbf{0.7945} & \cellcolor{OursRow}\textbf{0.7918} & \cellcolor{OursRow}\underline{0.9300} & \cellcolor{OursRow}\textbf{0.9420} & \cellcolor{OursRow}\underline{0.8185} & \cellcolor{OursRow}0.8082 & \cellcolor{OursRow}\underline{0.8655} & \cellcolor{OursRow}\textbf{0.8650} \\
\midrule
\multirow{7}{*}{\textbf{Iter.\ 2}} & \cellcolor{AllRow}All & \multicolumn{2}{c}{\cellcolor{AllRow}0.9164} & \multicolumn{2}{c}{\cellcolor{AllRow}0.7852} & \multicolumn{2}{c}{\cellcolor{AllRow}0.9160} & \multicolumn{2}{c}{\cellcolor{AllRow}0.7979} & \multicolumn{2}{c}{\cellcolor{AllRow}0.8539} \\
 & \cellcolor{BaselineRow}Random & \cellcolor{BaselineRow}0.9147 & \cellcolor{BaselineRow}\underline{0.9181} & \cellcolor{BaselineRow}0.7918 & \cellcolor{BaselineRow}0.7866 & \cellcolor{BaselineRow}\underline{0.9440} & \cellcolor{BaselineRow}0.9320 & \cellcolor{BaselineRow}0.8014 & \cellcolor{BaselineRow}0.8048 & \cellcolor{BaselineRow}0.8630 & \cellcolor{BaselineRow}0.8604 \\
 & \cellcolor{BaselineRow}Attribution & \cellcolor{BaselineRow}0.9113 & \cellcolor{BaselineRow}0.9172 & \cellcolor{BaselineRow}0.7907 & \cellcolor{BaselineRow}0.7889 & \cellcolor{BaselineRow}0.9300 & \cellcolor{BaselineRow}\underline{0.9340} & \cellcolor{BaselineRow}\textbf{0.8253} & \cellcolor{BaselineRow}0.7945 & \cellcolor{BaselineRow}\underline{0.8643} & \cellcolor{BaselineRow}0.8587 \\
 & \cellcolor{BaselineRow}Diversity & \cellcolor{BaselineRow}0.9121 & \cellcolor{BaselineRow}0.9147 & \cellcolor{BaselineRow}0.7915 & \cellcolor{BaselineRow}\underline{0.7903} & \cellcolor{BaselineRow}0.9180 & \cellcolor{BaselineRow}\underline{0.9340} & \cellcolor{BaselineRow}\underline{0.8185} & \cellcolor{BaselineRow}\underline{0.8082} & \cellcolor{BaselineRow}0.8600 & \cellcolor{BaselineRow}\underline{0.8618} \\
 & \cellcolor{BaselineRow}Attr-Div & \cellcolor{BaselineRow}\textbf{0.9215} & \cellcolor{BaselineRow}\textbf{0.9198} & \cellcolor{BaselineRow}\textbf{0.7928} & \cellcolor{BaselineRow}0.7852 & \cellcolor{BaselineRow}0.9220 & \cellcolor{BaselineRow}0.9240 & \cellcolor{BaselineRow}0.7945 & \cellcolor{BaselineRow}0.7979 & \cellcolor{BaselineRow}0.8577 & \cellcolor{BaselineRow}0.8567 \\
  & \cellcolor{BaselineRow}TSDS & \cellcolor{BaselineRow}\underline{0.9198} & \cellcolor{BaselineRow}0.9181 & \cellcolor{BaselineRow}0.7913 & \cellcolor{BaselineRow}0.7912 & \cellcolor{BaselineRow}0.9240 & \cellcolor{BaselineRow}\textbf{0.9440} & \cellcolor{BaselineRow}0.8151 & \cellcolor{BaselineRow}0.8116 & \cellcolor{BaselineRow}0.8625 & \cellcolor{BaselineRow}0.8662 \\
 & \cellcolor{OursRow}\textbf{\method} & \cellcolor{OursRow}\underline{0.9198} & \cellcolor{OursRow}\textbf{0.9198} & \cellcolor{OursRow}\underline{0.7922} & \cellcolor{OursRow}\textbf{0.7914} & \cellcolor{OursRow}\textbf{0.9460} & \cellcolor{OursRow}\textbf{0.9440} & \cellcolor{OursRow}0.8082 & \cellcolor{OursRow}\textbf{0.8151} & \cellcolor{OursRow}\textbf{0.8665} & \cellcolor{OursRow}\textbf{0.8676} \\
\midrule
\bottomrule
\end{tabular}
}%
\caption{Performance comparison on scientific datasets (\texttt{14b} base model). \textbf{Bold} and \underline{underscore} denote top-1/2 accuracy.}
\label{tab:weak-table-1}
\end{table*}

\begin{table*}[!htbp]
\centering
\renewcommand{\arraystretch}{0.75}
\resizebox{1.0\textwidth}{!}{%
\begin{tabular}{c l cc cc cc cc}
\toprule
\midrule
 &  & \multicolumn{2}{c}{\textbf{CommonsenseQA}} & \multicolumn{2}{c}{\textbf{LogiQA}} & \multicolumn{2}{c}{\textbf{LogiQA2}} & \multicolumn{2}{c}{\textbf{\textbf{Avg.}}} \\
\cmidrule(lr){3-4} \cmidrule(lr){5-6} \cmidrule(lr){7-8} \cmidrule(lr){9-10}
 & \textbf{Selection Ratio} $\rightarrow$ & \multicolumn{1}{c}{0.2} & \multicolumn{1}{c}{0.5} & \multicolumn{1}{c}{0.2} & \multicolumn{1}{c}{0.5} & \multicolumn{1}{c}{0.2} & \multicolumn{1}{c}{0.5} & \multicolumn{1}{c}{0.2} & \multicolumn{1}{c}{0.5} \\
\cmidrule{2-10}
 & \textbf{Methods} $\downarrow$ & & & & & & & & \\
\cmidrule{2-10}
 & \cellcolor{BaseRow}Base & \multicolumn{2}{c}{\cellcolor{BaseRow}0.8313} & \multicolumn{2}{c}{\cellcolor{BaseRow}0.5853} & \multicolumn{2}{c}{\cellcolor{BaseRow}0.4898} & \multicolumn{2}{c}{\cellcolor{BaseRow}0.6355} \\
\midrule
\multirow{7}{*}{\textbf{Iter.\ 1}} & \cellcolor{AllRow}All & \multicolumn{2}{c}{\cellcolor{AllRow}0.8477} & \multicolumn{2}{c}{\cellcolor{AllRow}0.5822} & \multicolumn{2}{c}{\cellcolor{AllRow}0.4975} & \multicolumn{2}{c}{\cellcolor{AllRow}0.6425} \\
 & \cellcolor{BaselineRow}Random & \cellcolor{BaselineRow}0.8460 & \cellcolor{BaselineRow}0.8468 & \cellcolor{BaselineRow}\underline{0.5945} & \cellcolor{BaselineRow}\underline{0.5914} & \cellcolor{BaselineRow}0.5000 & \cellcolor{BaselineRow}0.5019 & \cellcolor{BaselineRow}\underline{0.6468} & \cellcolor{BaselineRow}0.6467 \\
 & \cellcolor{BaselineRow}Attribution & \cellcolor{BaselineRow}\textbf{0.8518} & \cellcolor{BaselineRow}0.8509 & \cellcolor{BaselineRow}0.5822 & \cellcolor{BaselineRow}0.5730 & \cellcolor{BaselineRow}0.5038 & \cellcolor{BaselineRow}0.5038 & \cellcolor{BaselineRow}0.6459 & \cellcolor{BaselineRow}0.6426 \\
 & \cellcolor{BaselineRow}Diversity & \cellcolor{BaselineRow}0.8354 & \cellcolor{BaselineRow}\underline{0.8542} & \cellcolor{BaselineRow}0.5883 & \cellcolor{BaselineRow}0.5745 & \cellcolor{BaselineRow}\textbf{0.5159} & \cellcolor{BaselineRow}\textbf{0.5140} & \cellcolor{BaselineRow}0.6465 & \cellcolor{BaselineRow}\underline{0.6476} \\
 & \cellcolor{BaselineRow}Attr-Div & \cellcolor{BaselineRow}\underline{0.8493} & \cellcolor{BaselineRow}0.8444 & \cellcolor{BaselineRow}0.5684 & \cellcolor{BaselineRow}0.5745 & \cellcolor{BaselineRow}\underline{0.5153} & \cellcolor{BaselineRow}0.5064 & \cellcolor{BaselineRow}0.6443 & \cellcolor{BaselineRow}0.6418 \\
  & \cellcolor{BaselineRow}TSDS & \cellcolor{BaselineRow}0.8444 & \cellcolor{BaselineRow}0.8370 & \cellcolor{BaselineRow}0.5883 & \cellcolor{BaselineRow}0.5837 & \cellcolor{BaselineRow}0.5102 & \cellcolor{BaselineRow}0.4943 & \cellcolor{BaselineRow}0.6476 & \cellcolor{BaselineRow}0.6383 \\
 & \cellcolor{OursRow}\textbf{\method} & \cellcolor{OursRow}0.8452 & \cellcolor{OursRow}\textbf{0.8559} & \cellcolor{OursRow}\textbf{0.5960} & \cellcolor{OursRow}\textbf{0.5975} & \cellcolor{OursRow}0.5095 & \cellcolor{OursRow}\underline{0.5102} & \cellcolor{OursRow}\textbf{0.6502} & \cellcolor{OursRow}\textbf{0.6545} \\
\midrule
\multirow{7}{*}{\textbf{Iter.\ 2}} & \cellcolor{AllRow}All & \multicolumn{2}{c}{\cellcolor{AllRow}0.8411} & \multicolumn{2}{c}{\cellcolor{AllRow}0.5668} & \multicolumn{2}{c}{\cellcolor{AllRow}0.5032} & \multicolumn{2}{c}{\cellcolor{AllRow}0.6370} \\
 & \cellcolor{BaselineRow}Random & \cellcolor{BaselineRow}0.8452 & \cellcolor{BaselineRow}0.8509 & \cellcolor{BaselineRow}0.5730 & \cellcolor{BaselineRow}\underline{0.5899} & \cellcolor{BaselineRow}0.5115 & \cellcolor{BaselineRow}0.4987 & \cellcolor{BaselineRow}0.6432 & \cellcolor{BaselineRow}\underline{0.6465} \\
 & \cellcolor{BaselineRow}Attribution & \cellcolor{BaselineRow}\textbf{0.8534} & \cellcolor{BaselineRow}0.8468 & \cellcolor{BaselineRow}\underline{0.5853} & \cellcolor{BaselineRow}0.5822 & \cellcolor{BaselineRow}0.5032 & \cellcolor{BaselineRow}\textbf{0.5013} & \cellcolor{BaselineRow}0.6473 & \cellcolor{BaselineRow}0.6434 \\
 & \cellcolor{BaselineRow}Diversity & \cellcolor{BaselineRow}\underline{0.8509} & \cellcolor{BaselineRow}0.8436 & \cellcolor{BaselineRow}0.5791 & \cellcolor{BaselineRow}0.5822 & \cellcolor{BaselineRow}\textbf{0.5204} & \cellcolor{BaselineRow}\underline{0.5000} & \cellcolor{BaselineRow}\underline{0.6501} & \cellcolor{BaselineRow}0.6419 \\
 & \cellcolor{BaselineRow}Attr-Div & \cellcolor{BaselineRow}0.8419 & \cellcolor{BaselineRow}\underline{0.8526} & \cellcolor{BaselineRow}0.5684 & \cellcolor{BaselineRow}0.5837 & \cellcolor{BaselineRow}\underline{0.5184} & \cellcolor{BaselineRow}0.4847 & \cellcolor{BaselineRow}0.6429 & \cellcolor{BaselineRow}0.6403 \\
  & \cellcolor{BaselineRow}TSDS & \cellcolor{BaselineRow}0.8436 & \cellcolor{BaselineRow}0.8428 & \cellcolor{BaselineRow}0.5806 & \cellcolor{BaselineRow}0.5868 & \cellcolor{BaselineRow}0.5070 & \cellcolor{BaselineRow}0.5000 & \cellcolor{BaselineRow}0.6437 & \cellcolor{BaselineRow}0.6432 \\
 & \cellcolor{OursRow}\textbf{\method} & \cellcolor{OursRow}0.8493 & \cellcolor{OursRow}\textbf{0.8559} & \cellcolor{OursRow}\textbf{0.6006} & \cellcolor{OursRow}\textbf{0.6022} & \cellcolor{OursRow}\underline{0.5184} & \cellcolor{OursRow}\textbf{0.5013} & \cellcolor{OursRow}\textbf{0.6561} & \cellcolor{OursRow}\textbf{0.6531} \\
\midrule
\bottomrule
\end{tabular}
}%
\caption{Performance comparison on commonsense/logical datasets (\texttt{14b} base model). \textbf{Bold} and \underline{underscore} denote top-1/2 accuracy.}
\label{tab:weak-table-2}
\end{table*}

\begin{table*}[!htbp]
\centering
\renewcommand{\arraystretch}{0.75}
\resizebox{1.0\textwidth}{!}{%
\begin{tabular}{c l cc cc cc cc}
\toprule
\midrule
 &  & \multicolumn{2}{c}{\textbf{Med-MCQA}} & \multicolumn{2}{c}{\textbf{MedQA}} & \multicolumn{2}{c}{\textbf{HeadQA}} & \multicolumn{2}{c}{\textbf{\textbf{Avg.}}} \\
\cmidrule(lr){3-4} \cmidrule(lr){5-6} \cmidrule(lr){7-8} \cmidrule(lr){9-10}
 & \textbf{Selection Ratio} $\rightarrow$ & \multicolumn{1}{c}{0.2} & \multicolumn{1}{c}{0.5} & \multicolumn{1}{c}{0.2} & \multicolumn{1}{c}{0.5} & \multicolumn{1}{c}{0.2} & \multicolumn{1}{c}{0.5} & \multicolumn{1}{c}{0.2} & \multicolumn{1}{c}{0.5} \\
\cmidrule{2-10}
 & \textbf{Methods} $\downarrow$ & & & & & & & & \\
\cmidrule{2-10}
 & \cellcolor{BaseRow}Base & \multicolumn{2}{c}{\cellcolor{BaseRow}0.6075} & \multicolumn{2}{c}{\cellcolor{BaseRow}0.7023} & \multicolumn{2}{c}{\cellcolor{BaseRow}0.7801} & \multicolumn{2}{c}{\cellcolor{BaseRow}0.6966} \\
\midrule
\multirow{7}{*}{\textbf{Iter.\ 1}} & \cellcolor{AllRow}All & \multicolumn{2}{c}{\cellcolor{AllRow}0.6139} & \multicolumn{2}{c}{\cellcolor{AllRow}0.7172} & \multicolumn{2}{c}{\cellcolor{AllRow}0.8023} & \multicolumn{2}{c}{\cellcolor{AllRow}0.7111} \\
 & \cellcolor{BaselineRow}Random & \cellcolor{BaselineRow}\underline{0.6180} & \cellcolor{BaselineRow}0.6139 & \cellcolor{BaselineRow}0.7062 & \cellcolor{BaselineRow}0.7062 & \cellcolor{BaselineRow}0.7994 & \cellcolor{BaselineRow}0.7943 & \cellcolor{BaselineRow}0.7079 & \cellcolor{BaselineRow}0.7048 \\
 & \cellcolor{BaselineRow}Attribution & \cellcolor{BaselineRow}\textbf{0.6192} & \cellcolor{BaselineRow}0.6144 & \cellcolor{BaselineRow}0.7086 & \cellcolor{BaselineRow}\textbf{0.7141} & \cellcolor{BaselineRow}\textbf{0.8056} & \cellcolor{BaselineRow}\underline{0.7994} & \cellcolor{BaselineRow}\textbf{0.7111} & \cellcolor{BaselineRow}\underline{0.7093} \\
 & \cellcolor{BaselineRow}Diversity & \cellcolor{BaselineRow}0.6132 & \cellcolor{BaselineRow}0.6156 & \cellcolor{BaselineRow}0.7054 & \cellcolor{BaselineRow}0.6999 & \cellcolor{BaselineRow}0.8001 & \cellcolor{BaselineRow}0.7976 & \cellcolor{BaselineRow}0.7062 & \cellcolor{BaselineRow}0.7044 \\
 & \cellcolor{BaselineRow}Attr-Div & \cellcolor{BaselineRow}0.6158 & \cellcolor{BaselineRow}\textbf{0.6180} & \cellcolor{BaselineRow}0.7054 & \cellcolor{BaselineRow}0.7038 & \cellcolor{BaselineRow}0.7980 & \cellcolor{BaselineRow}0.7987 & \cellcolor{BaselineRow}0.7064 & \cellcolor{BaselineRow}0.7068 \\
  & \cellcolor{BaselineRow}TSDS & \cellcolor{BaselineRow}0.6101 & \cellcolor{BaselineRow}0.6139 & \cellcolor{BaselineRow}\textbf{0.7135} & \cellcolor{BaselineRow}0.6991 & \cellcolor{BaselineRow}0.7947 & \cellcolor{BaselineRow}0.7965 & \cellcolor{BaselineRow}0.7061 & \cellcolor{BaselineRow}0.7031 \\
 & \cellcolor{OursRow}\textbf{\method} & \cellcolor{OursRow}0.6120 & \cellcolor{OursRow}\underline{0.6173} & \cellcolor{OursRow}\underline{0.7101} & \cellcolor{OursRow}\underline{0.7109} & \cellcolor{OursRow}\underline{0.8020} & \cellcolor{OursRow}\textbf{0.8005} & \cellcolor{OursRow}\underline{0.7080} & \cellcolor{OursRow}\textbf{0.7096} \\
\midrule
\multirow{7}{*}{\textbf{Iter.\ 2}} & \cellcolor{AllRow}All & \multicolumn{2}{c}{\cellcolor{AllRow}0.6158} & \multicolumn{2}{c}{\cellcolor{AllRow}0.7093} & \multicolumn{2}{c}{\cellcolor{AllRow}0.7950} & \multicolumn{2}{c}{\cellcolor{AllRow}0.7067} \\
 & \cellcolor{BaselineRow}Random & \cellcolor{BaselineRow}0.6132 & \cellcolor{BaselineRow}0.6132 & \cellcolor{BaselineRow}\underline{0.7133} & \cellcolor{BaselineRow}0.7062 & \cellcolor{BaselineRow}0.7958 & \cellcolor{BaselineRow}0.7969 & \cellcolor{BaselineRow}0.7074 & \cellcolor{BaselineRow}0.7054 \\
 & \cellcolor{BaselineRow}Attribution & \cellcolor{BaselineRow}\underline{0.6146} & \cellcolor{BaselineRow}0.6153 & \cellcolor{BaselineRow}0.7086 & \cellcolor{BaselineRow}0.7031 & \cellcolor{BaselineRow}0.7969 & \cellcolor{BaselineRow}0.7994 & \cellcolor{BaselineRow}0.7067 & \cellcolor{BaselineRow}0.7059 \\
 & \cellcolor{BaselineRow}Diversity & \cellcolor{BaselineRow}0.6122 & \cellcolor{BaselineRow}0.6144 & \cellcolor{BaselineRow}0.7054 & \cellcolor{BaselineRow}0.7031 & \cellcolor{BaselineRow}\textbf{0.8038} & \cellcolor{BaselineRow}\textbf{0.8045} & \cellcolor{BaselineRow}0.7071 & \cellcolor{BaselineRow}\underline{0.7073} \\
 & \cellcolor{BaselineRow}Attr-Div & \cellcolor{BaselineRow}0.6122 & \cellcolor{BaselineRow}\underline{0.6158} & \cellcolor{BaselineRow}0.7101 & \cellcolor{BaselineRow}\underline{0.7078} & \cellcolor{BaselineRow}\underline{0.8005} & \cellcolor{BaselineRow}0.7976 & \cellcolor{BaselineRow}\underline{0.7076} & \cellcolor{BaselineRow}0.7071 \\
  & \cellcolor{BaselineRow}TSDS & \cellcolor{BaselineRow}0.6034 & \cellcolor{BaselineRow}0.6113& \cellcolor{BaselineRow}0.7038 & \cellcolor{BaselineRow}0.6968 & \cellcolor{BaselineRow}0.7991 & \cellcolor{BaselineRow}0.7950 & \cellcolor{BaselineRow}0.7021 & \cellcolor{BaselineRow}0.7010 \\
 & \cellcolor{OursRow}\textbf{\method} & \cellcolor{OursRow}\textbf{0.6158} & \cellcolor{OursRow}\textbf{0.6187} & \cellcolor{OursRow}\textbf{0.7164} & \cellcolor{OursRow}\textbf{0.7086} & \cellcolor{OursRow}0.7998 & \cellcolor{OursRow}\underline{0.8016} & \cellcolor{OursRow}\textbf{0.7107} & \cellcolor{OursRow}\textbf{0.7096} \\
\midrule
\bottomrule
\end{tabular}
}%
\caption{Performance comparison on biomedical/healthcare datasets (\texttt{14b} base model). \textbf{Bold} and \underline{underscore} denote top-1/2 accuracy.}
\label{tab:weak-table-3}
\end{table*}